\documentclass[12pt]{article} 

\pdfoutput=1

\usepackage{mathtools}
\usepackage{amssymb}
\usepackage{newtxtext,newtxmath} 
\usepackage{subcaption}
\usepackage{dsfont}
\usepackage{booktabs}
\usepackage{multirow}
\usepackage{bm}
\usepackage[noadjust]{cite}
\usepackage[top=2cm, bottom = 2cm, left= 1.5cm, right = 1.5cm]{geometry}
\usepackage[absolute]{textpos}
\usepackage{xcolor}

\DeclarePairedDelimiter{\norm}{\lVert}{\rVert}

\begin{document}
\title{Data-Driven Convergence Prediction of Astrobots Swarms\footnote{This work was financially supported by the Swiss National Science Foundation (SNF) Grant No. 20FL21\_185771 and the SLOAN ARC/EPFL Agreement No. SSP523.}}
\author{Matin~Macktoobian$^{a}\footnote{matin.macktoobian@epfl.ch}$, Francesco~Basciani$^{b}$, Denis~Gillet$^{a}$, and Jean-Paul~Kneib$^{a}$}
\date{%
	$^a$EPFL, Lausanne, Switzerland\\%
	$^b$ Turin Polytechnic, Turin, Italy\\%
}


\maketitle

\begin{abstract}
Astrobots are robotic artifacts whose swarms are used in astrophysical studies to generate the map of the observable universe. These swarms have to be coordinated with respect to various desired observations. Such coordination are so complicated that distributed swarm controllers cannot always coordinate enough astrobots to fulfill the minimum data desired to be obtained in the course of observations. Thus, a convergence verification is necessary to check the suitability of a coordination before its execution. However, a formal verification method does not exist for this purpose. In this paper, we instead use machine learning to predict the convergence of astrobots swarm. In particular, we propose a weighted $k$-NN-based algorithm which requires the initial status of a swarm as well as its observational targets to predict its convergence. Our algorithm learns to predict based on the coordination data obtained from previous coordination of the desired swarm. This method first generates a convergence probability for each astrobot based on a distance metric. Then, these probabilities are transformed to either a complete or an incomplete categorical result. The method is applied to two typical swarms including 116 and 487 astrobots. It turns out that the correct prediction of successful coordination may be up to 80\% of overall predictions. Thus, these results witness the efficient accuracy of our predictive convergence analysis strategy.   
\end{abstract}
\def\abstractname{Note to Practitioners}
\begin{abstract}
Observatories involved in the generation of spectroscopic surveys always encounter limited resources to check the throughputs of their planned observations before their executions. The information yielded by an observation directly depend on the convergence rate of the observatory’s astrobots in that particular observation. Namely, if the astrobots’ convergence rate is below a minimum, then the observation has to be revoked and re-planned. So, one may define another observation which fulfills the minimum-information requirement. There has been yet no analytical tool developed to verify the convergence rate of the coordination computed by the state-of-the-art trajectory planners of astrobots swarms. Thus, we propose to use a machine learning scheme to predict the desired convergence rate instead of involving in the infeasible process of finding its exact value. This method is a supervised method which requires the target-to-astrobot assignments table of an observation. The algorithm also needs a dataset including previous coordination results of various observations of a particular swarm. The simulated scenarios manifest magnificent accuracies in the convergence predictions of the some astrobots swarms corresponding to modern spectroscopic surveys such as SDSS-V (including $\sim$500 astrobots). Our strategy is based on the smallest subset of the astrobots' features which have a pivotal role in convergence rates, say, the projected positions of targets on a hosting telescope’s focal plane. We argue that more explorations have to be considered to find other important features, such as the motion direction of each astrobot, which may even further improve the obtained prediction accuracies.
\end{abstract}
\textbf{keywords}: astrobotics, convergence prediction, machine learning, swarm robotics, spectroscopic surveys, astronomical instrumentation
\section{Introduction\protect\footnote{Throughout this paper, scalars and (sets of) matrices are represented by regular and bold symbols, respectively.}}
The unknown nature of dark matter and dark energy is among the most major gaps in the modern physics \cite{arkani2009theory}. Cosmology have actively sought the history of the universe, which is known to be tied with the evolution of dark matter. A unified mathematical model of dark matter has not yet been achieved based on analytical methods \cite{jungman1996supersymmetric}. Thus, cosmologists have shifted their attention to observational data in various red shift ranges \cite{zhao2017dynamical,newman2015spectroscopic}. Each range of redshift represents a particular time interval corresponding to the universe's lifespan. So, the recent trend in dark matter studies aims to generate the map of the observable universe. Then, the analysis of such a map would eventually reveal new findings about the distribution of dark matter all over the cosmos. Cosmological spectroscopy is the front-runner technique to contribute to the cited goal. In particular, dominant massive objects of the universe, say, galaxies, quasars, etc., all emanate electromagnetic radiations. These radiations can be captured in particular wavelengths by optical fibers mounted on specific ground telescopes. For this purpose, many optical fibers are placed at a particular area of a candidate ground telescope which is called focal plane. The generation of the map of the observable universe is not a trivial task given the huge number of the target objects residing in it. So, a set of observations are defined each of which includes a subset of the all those targets. In this regard, the local map of each observation is a survey. So, the eventual accumulation of many surveys gives rise the complete map of the observable universe. To observe those objects, i.e., capturing their light, each target has to be assigned to one of the optical fibers of a telescope. Then in the course of an observation's exposure time, the desired rays are collected by the optical fibers. Later, a spectrograph connected to the optical fibers synthesizes the spectroscopic survey corresponding to the planned observation. The number of spectroscopic survey projects has been increased during of the recent decade the most prominent of which are DESI \cite{flaugher2014dark}, MOONS \cite{cirasuolo2014moons}, PFS \cite{ellis2012extragalactic}, SDSS-V \cite{kollmeier2017sdss}, LSST \cite{mandelbaum2019wide}, MegaMapper \cite{schlegel2019astro2020}, etc.

Each observation comprises a unique set of targets. The location of each target obviously differs from those of other targets. Thus from one observation to another, one has to change the configuration of the fibers so that tip of each fiber points its new target associated with a new observation. In the first generation of spectroscopic surveys, these coordination procedures were manually done using various techniques such as magnetic fiber technology \cite{lewis2014fibre,fabricant2005hectospec} and slit masks \cite{dressler2006imacs,mclean2010design}. However, such manual coordinations are proved to be inefficient in the case of the requirements of the recent advanced spectroscopic projects. First, the current projects are equipped with hundreds to thousands of fibers. The available time to coordinate fibers from one observation to another is limited. In this case, if the fibers are not coordinated on time, their observation's data would be partially collected in the best case. On the other hand, each observation depends on many celestial factors whose second fulfillment may require very long times. So, planned observations must not be missed according to survey programs. In other words, cosmologists need fast automatic coordination of fibers. Second, the more fibers one places in a focal plane, the larger surveys may be taken into account. Increasing the density of fiber placements makes manual coordination even more challenging because they may disturb the calibration of fibers' tips. This also magnifies the need to minimize human interventions in the coordination processes.

To resolve the issues above, the idea of astrobotics have been emerged. Each astrobot \cite{horler2018robotic} is a two-degree-of-freedom rotational-rotational manipulator which contains a fiber. To be specific, a fiber passes through the central axis of its astrobot so that the fiber's tip is located at the end-effector of the astrobot called ferrule. In this case, the tip of the fiber indeed can reach any point in the circular surface corresponding to the working space of its astrobot's ferrule. So given any target assigned to a fiber \cite{morales2011fibre,macktoobian2020optimal}, should the target reside in its astrobot's working space, the astrobot may be controlled so that its ferrule reaches the projected location of the target on the focal plane. As stated before, one intends to maximize the number of the fibers on a telescope. Thus, astrobots have to be placed, in hexagonal formations, so close to each other that their working spaces unavoidably overlap. These overlapping areas imply the possibility of collisions between various astrobots in the course of their coordination toward their targets. So, the coordination problem of astrobots swarms is inherently safety-critical for which various control strategies were proposed. For example, nonlinear hybrid control was taken into account \cite{makarem2016collision,tao2018priority} to realize not only collision avoidance but also coordination priority for the astrobots whose targets are more important than those of other targets in view of the signals they collect. This method cannot generally coordinate all astrobots, so the observational information reflected into surveys are not maximized. In this regard, the formulation of nonlinear hybrid control was revised \cite{macktoobian2019navigation, macktoobian2019complete} so that one can check whether or not a particular setup of astrobots can be totally converged to their targets. This method, despite of its merit in completeness determination, is computationally so expensive that its real-time application may not be always feasible if the available times between successive coordination are too short. The convergence rate assessment of coordination may be done using numerical simulations of coordination with respect to various observation settings. This procedure is useful for small and medium surveys but not massive ones. Namely, convergence rate assessment requires the real-time solutions of hundreds to thousands of interdependent differential equations corresponding to distributed navigation functions of astrobots. Such analyses may not be feasible in the case of tight observation schedules in which the available times between observations are not long enough. If such assessment is possible, then inefficient coordination can be re-planned to those whose information throughput satisfy surveys expectations. In particular, a coordination output directly depends on the target-to-astrobot assignments corresponding to its observation. One may revisit an assignment to yield better coordination, thereby achieving higher convergence rates. Supervisory control was also employed to synthesize control commands whose safety and completeness can be formally verified \cite{macktoobian2019supervisory}. However, this strategy also becomes inefficient because of the curse of dimensionality in the case of crowded astrobots swarm.
\subsection{Literature Review}
Machine learning techniques have been partially contributed to the trajectory planning of multi-agent systems. For example in \cite{su2011using}, an anomaly network traffic identification problem is studied for autonomous vehicles. This problem conceptually resembles the collision avoidance aspect of our prediction problem. In this method, the overall working space of the problem is so vast, yet the number of the number of vehicles are relatively small. So, collision avoidance is not a critical issue in the assumed sparse distribution of vehicles. In contrast, our convergence prediction problem indeed implies hazardous interactions in dense formations of astrobots, thereby entailing considerable risk of collisions between them. Additionally, our convergence prediction problem also features noticeable sensitivity to even trivial spatial deviations of configurations in terms of convergence results. On that account, any potential dataset representing our problem needs to encompass sufficient data to cover a wide range of similar configurations. A similar study takes the idea of moving ranges into account to assess neighbors more effectively for the vehicles in crowded urban areas \cite{lee2015moving}. The predictive model generated by this scheme relaxes the structured assumption by allowing movements of uncertain objects. The aforesaid relaxation, though, complicates the compliance with the safety requirements of this scenario. Instead, our problem enjoys the fully structured dynamics of astrobots swarm. Namely, the extremely constrained dynamics of each astrobot does not exert any uncertain feature to the prediction problem. As another example, learning-based strategies have been employed to predict trajectories of multi-agent systems in unconstrained or loosely constrained systems. For instance, route prediction for ships was investigated \cite{duca2017k}. This study uses a variation of $k$-NN algorithm which exclusively models each ship as an isolated entity, say, in the absence of any collisions with other peers. 

Limited applications of machine learning in trajectory prediction of more complex swarms are also reported. To give an instance, a class of aggregating behaviors in a self-organizing swarm were the subject of a prediction problem \cite{khaldi2018self} using distance-weighted $k$-NN method \cite{jin2019improvement,cataloluk2012diagnostic,liu2011class,gou2012new}. The density metric of the swarm is modeled by hydrodynamical particle interpolation. This system seeks predictions through fairly complicated movements scenarios. However, the goal is the classification of collective behaviors while the involved non-interacting agents are subject to no collisions. Collision freeness was interestingly taken into account in a coordination scenario using artificial potential fields \cite{chen2018collision}. This work is relatively comparable to what we seek in this paper, because the coordination control of astrobots is based on a class of artificial potential fields. However, the prediction application in this method is trivially concentrated on finding the closest point of an obstacle to a robot. Put differently, this strategy only guarantees collision freeness between a single robotic arm and a human's hand. Thus, in the absence of other agents, the complexity of this scenario is significantly less than what one encounters in the convergence prediction of astrobots. 
\begin{figure*}
	\centering
	\hspace*{-20mm}
	\begin{subfigure}[b]{.50\textwidth}
		\centering
		\includegraphics[width=\textwidth]{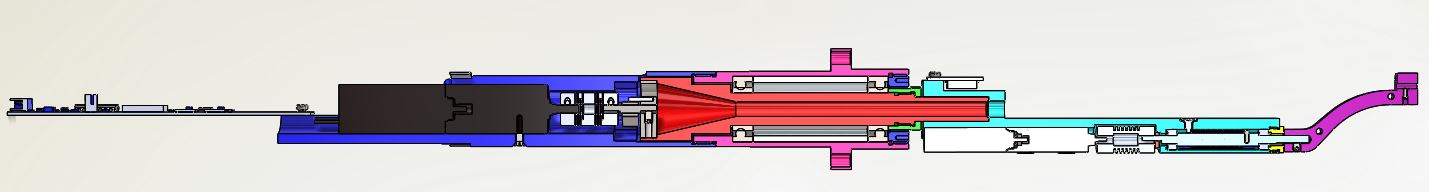}
		\caption{The side view of an astrobot\label{fig:pos}}%
		\vspace{-2mm}	
		\begin{minipage}[b]{.45\linewidth}
			\centering
			\subcaptionbox{The top view of an astrobot\label{fig:top}}
			{\includegraphics[scale=0.42]{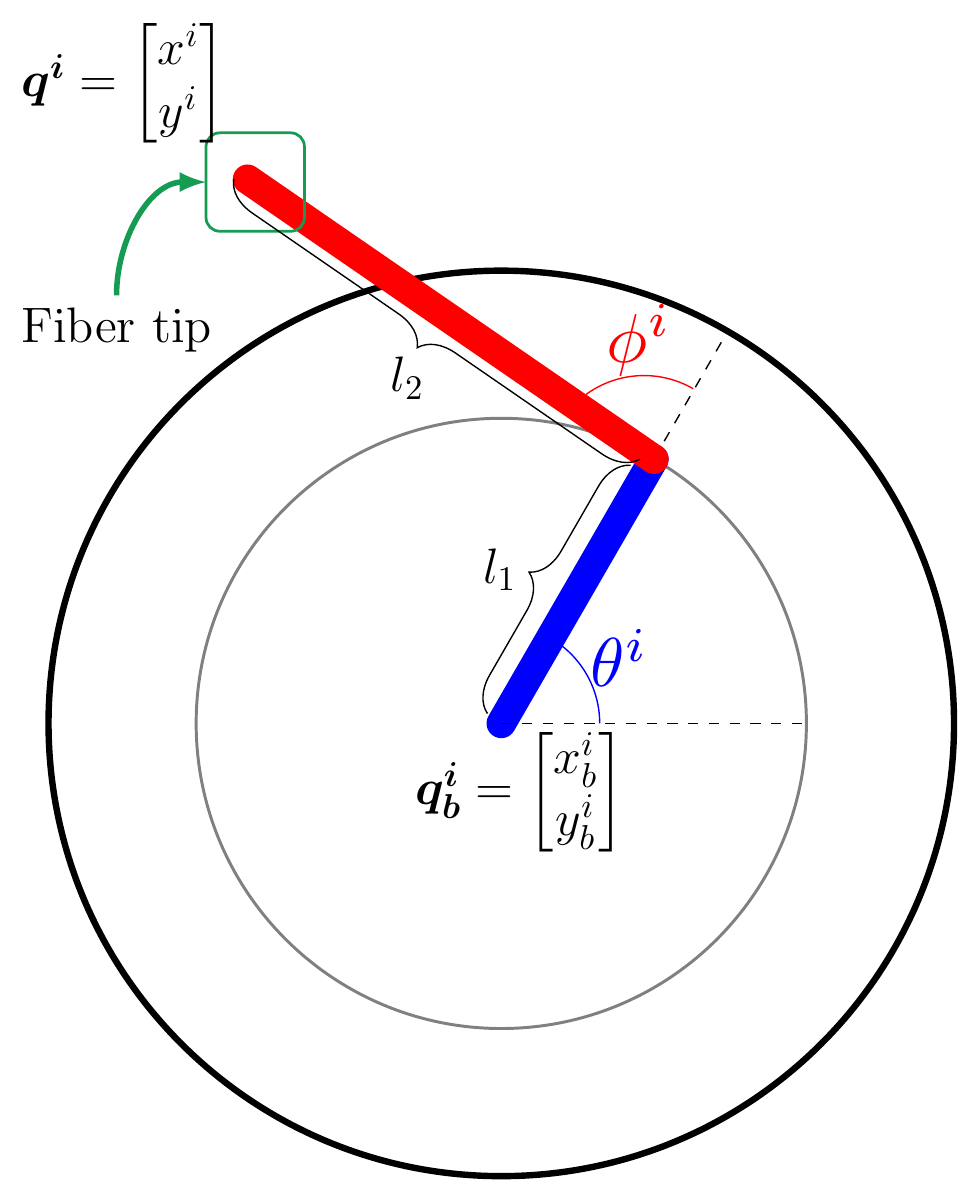}}
		\end{minipage}\quad
		\begin{minipage}[b]{.45\linewidth}
			\centering
			\subcaptionbox{A spectrograph\label{fig:spec}}
			{\includegraphics[scale=0.068]{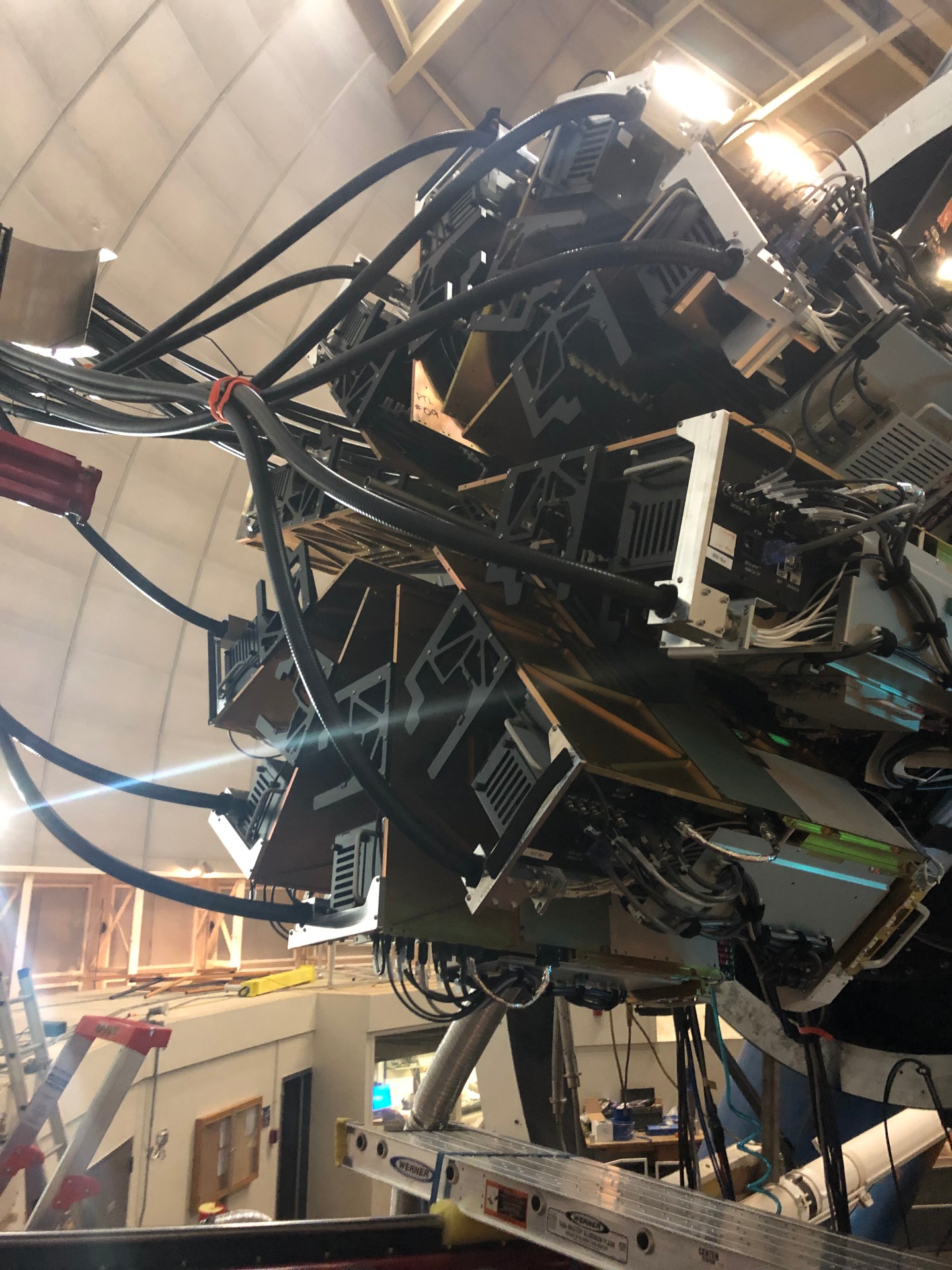}}
		\end{minipage}
	\end{subfigure}
	\quad
	\begin{subfigure}[b]{.4\textwidth}
		\centering
		\subcaptionbox{An astrobots swarm located at a telescope's focal plane\label{fig:swarm}}
		{\includegraphics[scale=0.13]{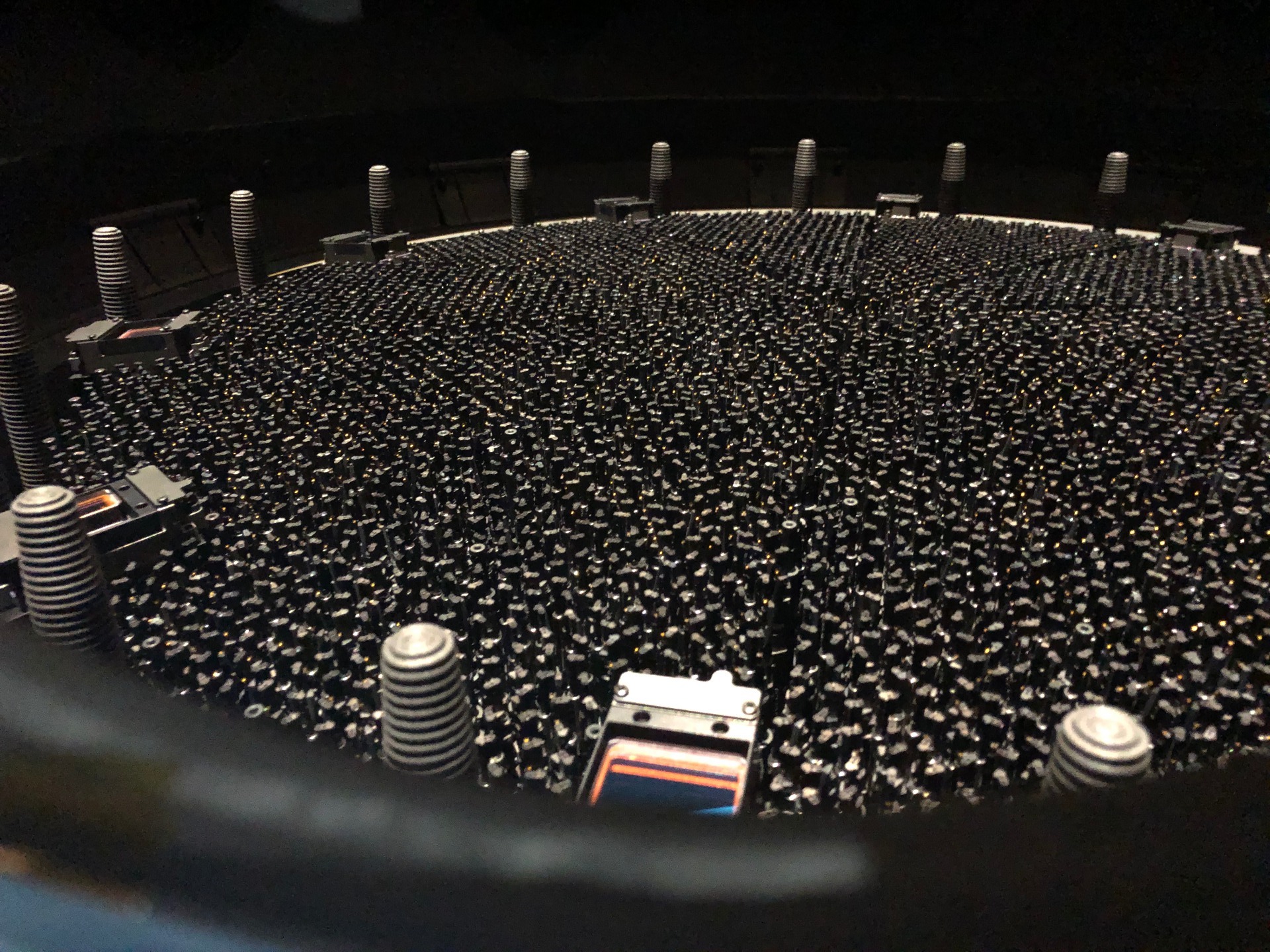}}
	\end{subfigure}
	\caption{An astrobots swarm and its elements (reprinted from \cite{horler2018robotic}, \cite{macktoobian2019navigation}, and DESI webpage in Ohio State University, respectively)}
\end{figure*}

The machine-learning-based behavioral predictions for multi-agent swarms have not been extensively studied. In particular, a learning system can efficiently train a model of a system if one feeds the data corresponding to all important features of that system. In the case of multi-agent swarms, these feature sets are often so large that final models may not be applicable for various reasons. First of all, training a predictive model requires enough data representing the behavioral patterns of system. The more complicated a system is, the more data of it one needs to effectively synthesize a predictor for it. The complexity of multi-agent swarms then requires huge datasets exhibiting their behaviors. But such amounts of data are often not available specially in the case of heterogeneous swarms. Moreover, a swarm system's functionalities are generally subject to many constraints whose presence may easily drive any learning model of that swarm toward common machine learning issues like underfitting and overfitting. Accordingly, the complete convergence of astrobots in the course of their coordination has not yet been efficiently resolved for the swarms including thousands of astrobots. On the other hand, partial coordination may lead to small convergence rates according to which the lack of enough data gives rise to the generation of the surveys whose wealth of information and details are not sufficient. Thus, instead of questing after analytical solutions to the completeness checking problem in more efficient ways, we shift our perspective to the prediction of complete coordination. In this framework, we seek to compute some models based on the data obtained from former coordination to predict the convergence rates of future ones in terms of some particular features. For this purpose, we propose a prediction algorithm based on the idea of weighted $k$-NN \cite{peterson2009k}, given the relative simplicity and design intuitions which stems from the geometrical formulation of $k$-NN-driven strategies. Subject to a set of astrobots assigned to their targets, our method predicts whether or not each astrobot would successfully converge to its target spot. The applied evaluations to simulated results using our scheme exhibit high performances in those predictions.
\subsection{Contributions}
We establish a predictive algorithm which paves the way for assessing the suitability of a particular astrobots-to-targets mapping set in terms of its expected information throughput. In other words, we propose a predictive solution to the decision making problem of whether a particular set of astrobot-to-target pairings would give rise to our expected number of successfully converged astrobots. This achievement is quite important if one takes the notion of observation priority in the definition of a survey plan. In particular, each survey plan may include some targets whose observations have more remarkable impact on the quality of the final survey. In this regard, a successful coordination may be defined as the one through which the astrobots corresponding to high-priority targets can be reached. Our algorithm individually predicts the convergence of each astrobot. Thus, the priority-based decision making process may also be covered using out method by exclusively focusing on the reachability prediction of high-priority targets.

A coordination process is a finite set of movements corresponding to each astrobot of a swarm with respect to many functional and safety requirements. A formal convergence verification tool has to check every single coordination step according to the control signals generated for each astrobot in every step. However, the discussion presented in the previous sections clarified that such exact approach to convergence analysis may be practically infeasible. Thus, among all steps of a coordination process, our algorithm merely works based on the first (i.e., initial) and the last (i.e., final) astrobots-targets configurations of the process. Another challenge raises from the imbalanced nature of the data in our problem. Namely, the convergence rate of large astrobots swarms generally varies between $65\%\sim85\%$ depending on their populations. Thus, the number of the astrobots which converge is noticeably larger than those which don't converge. So, the data are inherently imbalanced. It is widely observed that imbalance data may adversely impact the output of any naive machine learning algorithm which does not counteract against this issue. The applied simulations of our algorithm with respect to various populations of large astrobots manifest its effectiveness in terms of various performance measures.
\subsection{Paper Outline}
The remainder of the paper is structured as follows. Section \ref{sec:rev} present a brief review on astrobots characterization and the swarms constructed by them. The important elements which play crucial rules in modeling individual astrobots and their swarms through coordination are illustrated. We then shift our attention to the specify the convergence prediction problem in Section \ref{sec:ps}. We particularly focus on the features according to which a data-driven solution to the convergence prediction problem is indeed challenging. Section \ref{sec:cps} comprises a weighted $k$-NN-based solution to the cited problem. We then present detailed statistical analysis to express the credibility of our algorithm in Section \ref{sec:sim}. We indeed apply our algorithm to two complex instances of astrobots swarms which include 116 and 487 astrobots. In the end, Section \ref{sec:conc} reflects our conclusions and discusses potential search ideas to improve our results in future.
\section{A Review on Astrobots Swarms}
\label{sec:rev}
Each astrobot is a robotic manipulator with two degrees of freedom whose schematic is depicted in Fig. \ref{fig:pos}. It is an active placeholder for the fiber which is passed through its central axis. So, the astrobot has to move its end-effector, called ferrule, so that the fiber may reach any point corresponding to the working space of the astrobot. For this purpose, two rotational arms of the astrobot represent its two degrees of freedom as shown in Fig. \ref{fig:top}. The overall length of two arms is long enough to each the centroid of any neighboring astrobot. So, astrobots can overall reach the whole surface of the focal plane which is a particular area of the telescope at which astrobots are mounted. All fibers are connected to a spectrograph, see, Fig. \ref{fig:spec}, which is located at the back of the focal plane. The spectrograph processes the signals collected by fibers to generate the survey corresponding to each observation. 

Because of astronomical requirements, astrobots are densely located in hexagonal formations in their hosting focal plane as illustrated in Fig. \ref{fig:swarm}. Such placement paves the way for the proper functionality of each fiber in terms of focal plane coverage and signal capturing. However, many challenges rise in view of coordinating astrobots from their initial configuration to a desired one by a particular observation. First, the dense formation of astrobots severely makes them subject to collisions. Thus, a controller has to plan some trajectories for each astrobot by which its ferrule reaches it desired spot. It is practically observed that if astrobots starts their movements from an arrangement in which their distances from each other is maximum, the planned trajectories would be obtained more efficiently. Thus, to alleviate the trajectory planning complexities in these highly-dense systems, astrobots are always reconfigured to their folded formation in which $\theta = 0$ and $\phi = \pi$ as rendered in Fig. \ref{fig:full}.
\section{Problem Statement}
\label{sec:ps}
The more astrobots converge to their target spots, the more the throughout of the observation associated with the targets will be. The current trajectory planners are not always able to achieve desired high convergence rates \cite{makarem2016collision}. If a convergence rate is below a certain threshold, then its corresponding final survey will not represent the expected quality. Thus, one has to assess the performance of a potential coordination process in terms of its final convergence before its execution. The analytical \cite{macktoobian2019complete} and logical \cite{macktoobian2019supervisory} tools to verify the results before their execution are often computationally too expensive. In this regard, these methods may not be used in real-time scenarios when the time slots available between observations are too short. The cited tools analyze every coordination step to check the collision freeness of motions which eventually tend to final configurations of astrobots. However, in this research, we only take the initial configuration of astrobots and the locations of their targets into account. We intend to predict whether or not a particular number of astrobots completely converged in the course of an observation surpasses the minimum number of desired convergences. Then, if the predicted convergence rate is larger than the minimum expectation, then we decide to let the trajectory planner coordinate our swarm. Otherwise, we re-plan the unsatisfactory astrobot-to-target assignments to yield better combinations.

\begin{figure}
	\centering
	\hspace*{-10mm}
	\begin{subfigure}[b]{.3\textwidth}
		\subcaptionbox{The folded formation of astrobots representing their initial configuration\label{fig:full}}%
		{\includegraphics[scale=0.6]{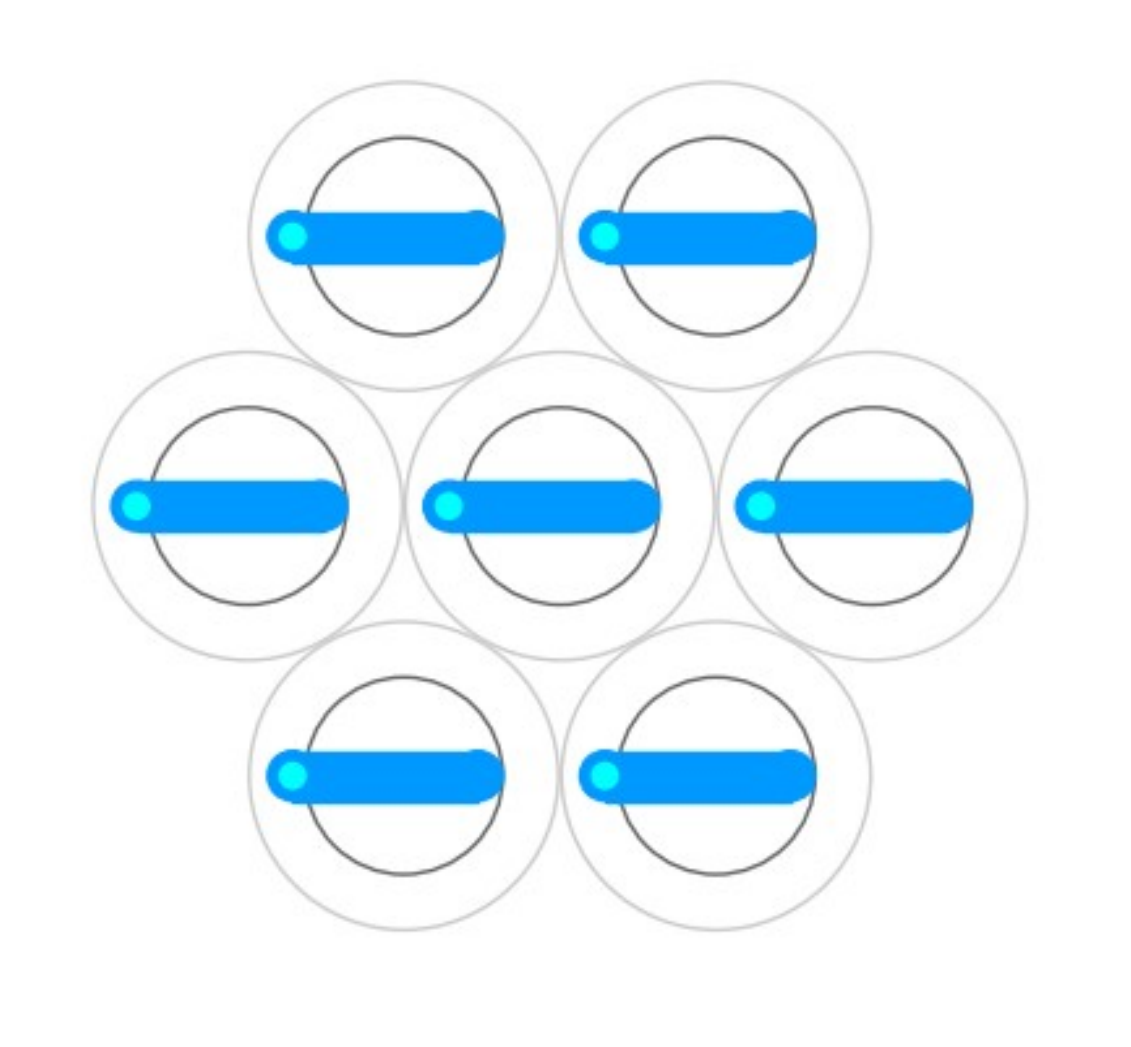}}
		\vspace{-0.9cm}	
	\end{subfigure}
	\quad
	\begin{subfigure}[b]{.3\textwidth}
		\hspace*{7mm}\subcaptionbox{The scheamtic of the convergence prodiction problem\label{fig:arch}}
		{\includegraphics[scale=0.9]{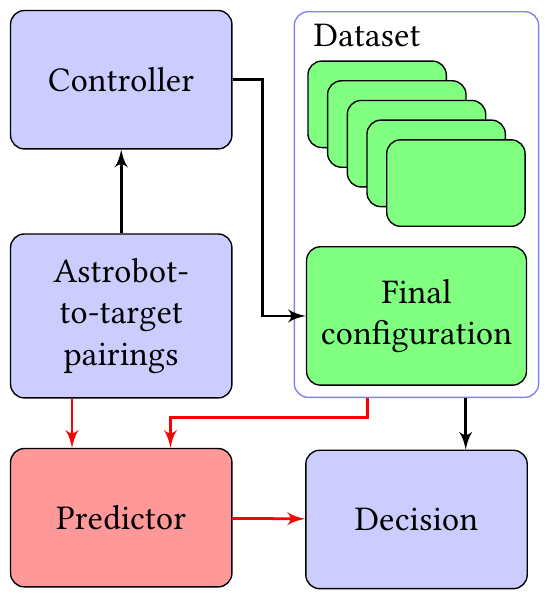}}
	\end{subfigure}
	\caption{From astrobots initial configurations to the convergence prediction problem statement}
\end{figure}

The problem statement is graphically shown in Fig. \ref{fig:arch} in which we seek the synthesis of a predictor to solve the problem. In particular, we prepare a dataset including many coordination scenarios with respect to multitude of astrobot-to-target assignment pairings which had been already simulated and/or executed. In this dataset, each astrobot in each pairing is labeled by 1 (resp., 0) if it finally reaches (resp., doesn't reach) its target. The overall set of this results is called \textit{ground truth vector}. We use these data to predict convergence rates using a weighted $k$-NN-based strategy. Since the number of converging astrobots is often larger than that of those which doesn't converge, our data are inherently biased. Such imbalance data have to become balanced to make predictions reliable. We also only consider safe coordination scenarios in our dataset.
\section{Convergence Prediction Strategy}
\label{sec:cps}
\begin{figure}
	\centering
	\includegraphics[scale=1]{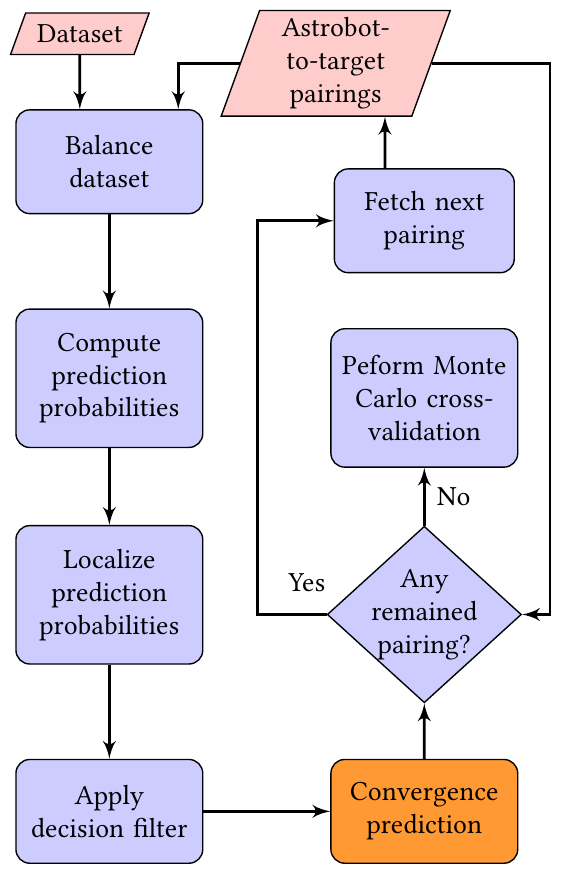}
	\caption{The flow of the convergence prediction algorithm}
	\label{fig:alg}
\end{figure}

In this section, we elaborate on our convergence prediction algorithm, as shown in Fig. \ref{fig:alg}. We first compensate the imbalanced data issue using a set of vector weights applied to our data. Then, a distance metric is defined to rank the astrobots neighborhoods with respect to a desired astrobot whose convergence is intended to be predicted. A prediction probability is computed associated with each astrobot. We then note that the prediction problem of each astrobot has to be essentially analyzed in its own neighborhood. Thus, we localize the analysis which is mathematically equivalent to a particular normalization of the quoted prediction probabilities. Next, given a particular \textit{decision filter}, we transform the obtained probabilities to either of two categorical outcomes. Each of these outcomes represents the prediction of our algorithm regrading the successful or the unsuccessful convergence of their corresponding astrobots. We finally perform Monte Carlo cross-validation \cite{xu2001monte} to assess the reliability of the results of our algorithm.

One notes that the coordinate associated with each astrobot's initial configuration is fixed (see, Fig. \ref{fig:full}), and it does not impact the coordination phase. Thus, in the prediction process, we define the astrobot vector $\bm{\pi}$ according to the location of its projected target on the focal plane of the swarm as follows\footnote{Unary operator $(\cdot)^\intercal$ represents the transpose of its matrix argument.}.
\begin{equation}
	\bm{\pi} \coloneqq \begin{bmatrix}
	x_{t} & y_{t}
	\end{bmatrix}^\intercal
\end{equation}
Then, the \textit{configuration matrix} $\bm{P}$ of a specific swarm $P$ including $n$ astrobots is indeed the accumulated configurations of its constituting astrobots which is
\begin{equation}
	\bm{P} \coloneqq \begin{bmatrix}
	x_{t^1} & x_{t^2} & \cdots & x_{t^n}\\
	y_{t^1}  &y_{t^2} & \cdots& y_{t^n}
	\end{bmatrix}^\intercal.
\end{equation}
The ground truth vector corresponding to $\bm{P}$ is $\bm{g^{P}}$. This vector represents the a posteriori information regarding the convergence of its corresponding configuration stored in a dataset. The more configurations exist in the dataset, the more representative the dataset is for its swarm. Since there are infinite number of configurations associated with a swarm, it is impossible to accumulate any possible coordination scenario in the dataset. However, the dataset has to be representative enough because changing the location of a target for just some tenths of millimeters just may change a successful convergence to a deadlock situation or vice versa. The dataset has to be divided into train and test partitions whose division proportion is discussed in Section \ref{sec:sim}.
\subsection{Imbalanced Data Compensation}
The family of $k$-NN algorithm is very sensitive to the local structure, i.e., the geometry, of data. We particularly enjoy this feature because the convergence prediction problem directly depends on geometrical characteristics of astrobot vectors. As already noted, configuration matrices often include many 1s compared to 0s because the majority of astrobots can be successfully coordinated using a swarm controller. So, their dataset is imbalanced according to which $k$-NN-based algorithms do not properly work \cite{dubey2013class}. There are two typical approaches to resolving this issue neither of which is effectively applicable to our case. In particular, one may perform an oversampling (resp., undersampling) on the minority class (resp., majority class). This approach is infeasible in our case because an oversampling on the minority class requires the configurations whose ground truth vectors have more 0s than 1s. In the case of huge swarms, such configurations are extremely rare, if not nonexistent. Even if one could find such configurations, the next step would be the generation of a new group of targets which are very close to the targets of that configuration. But, it would be so likely that many 1s are also generated, thereby essentially canceling the purpose of oversampling. On the other hand, any undersampling needs to remove all the configurations whose ground truth vectors include more 1s than 0s. However, it gives rise to the loss of valuable information which are important for potential prediction cases.

Instead, we devise a vector of weights to enhance the impact of  0s in the ground truth vector of a specific configuration. This strategy is similar to the idea of class confidence weights \cite{liu2011class}. The difference is that we apply the weights to single astrobots, not to data samples, i.e., configurations. Given a configuration $\bm{P}_{i}$ where $i \in \{1,2, \cdots, N\}$, assume that ground truth vector $\bm{g^{P}}_i$ is associated with it. We define \textit{frequency vector} $\bm{u}$ and its complement, say, \textit{pseudo vector} $\bm{v}$ as follows.
\begin{equation}
	\begin{split}
		\bm{u} &\coloneqq \sum_{i} \bm{g^\bm{P}}_i\\
		\bm{v} &\coloneqq N\cdot\mathds{1}_{1\times N} - \bm{u}
	\end{split}
\end{equation}
Then, the elements\footnote{$n$-ary operator $\bigcup\limits_{i}(\cdot)_{i}$ constructs a vector of the operator argument.} of \textit{weight vector} $\bm{w} \coloneqq \bigcup\limits_{i}w_{i}$ read as below.
\begin{equation}
	w_{i} \coloneqq 
	\begin{dcases*}
		u_{i} & if $v_{i} = 0$\\
		\dfrac{u_{i}}{v_{i}} & otherwise
	\end{dcases*}
\end{equation}
Each element of $\bm{w}$ has to be applied to the 0s of a particular astrobot of the configuration. We apply different weights to different astrobots because those which are in total neighbourhoods, i.e., surrounded by 6 astrobots, generally don’t reach their target positions as frequent as those which are in partial neighbourhoods configuration. So, the 0s of the astrobots in total neighbourhood configurations have smaller weights compared to those in partial neighborhoods. The notion of weight vector efficiently compensates the problem of imbalanced data. However in our problem, the two classes have not the same importance. In other words, we are more interested in the correct predictions of 1s rather than 0s in an operational point of view. So, we tune the elements of weight vectors according to our prediction requirements using two corrector coefficients $\alpha$ and $\beta$ on which we elaborate in Section \ref{sec:sim}.
\subsection{Prediction Probability Computation}
We define a distance metric to quantitatively compare various configurations with each other. Let $\bm{T}$ be a test configuration, say, the one we are interested in predicting its convergence. Let also $\bm{P}_{i}$ be a train configuration. We define \textit{distance metric} $\Delta(\cdot,\cdot)$ which later is used to find the close train configurations to a particular test one as below\footnote{Unary operator $\norm{\cdot}$ denotes the Euclidean norm of its vector argument.}
\begin{equation}
	\Delta(\bm{T},\bm{P}_{i}) \coloneqq \sum\limits_{j}\norm{\bm{T}_{j}-\bm{P}_{i,j}}.
\end{equation}
Here, $\bm{T}_{j}$ and $\bm{P}_{i,j}$ corresponds to the $j$th columns (i.e., astrobots) of $\bm{T}$ and $\bm{P}_{i}$, respectively. 
\begin{figure}
	\centering
	\hspace*{-10mm}
	\begin{subfigure}[b]{.35\textwidth}
		\hspace*{-2mm}\subcaptionbox{A hypothetical neighborhood\label{fig:neigh}}
		{\includegraphics[scale=0.6]{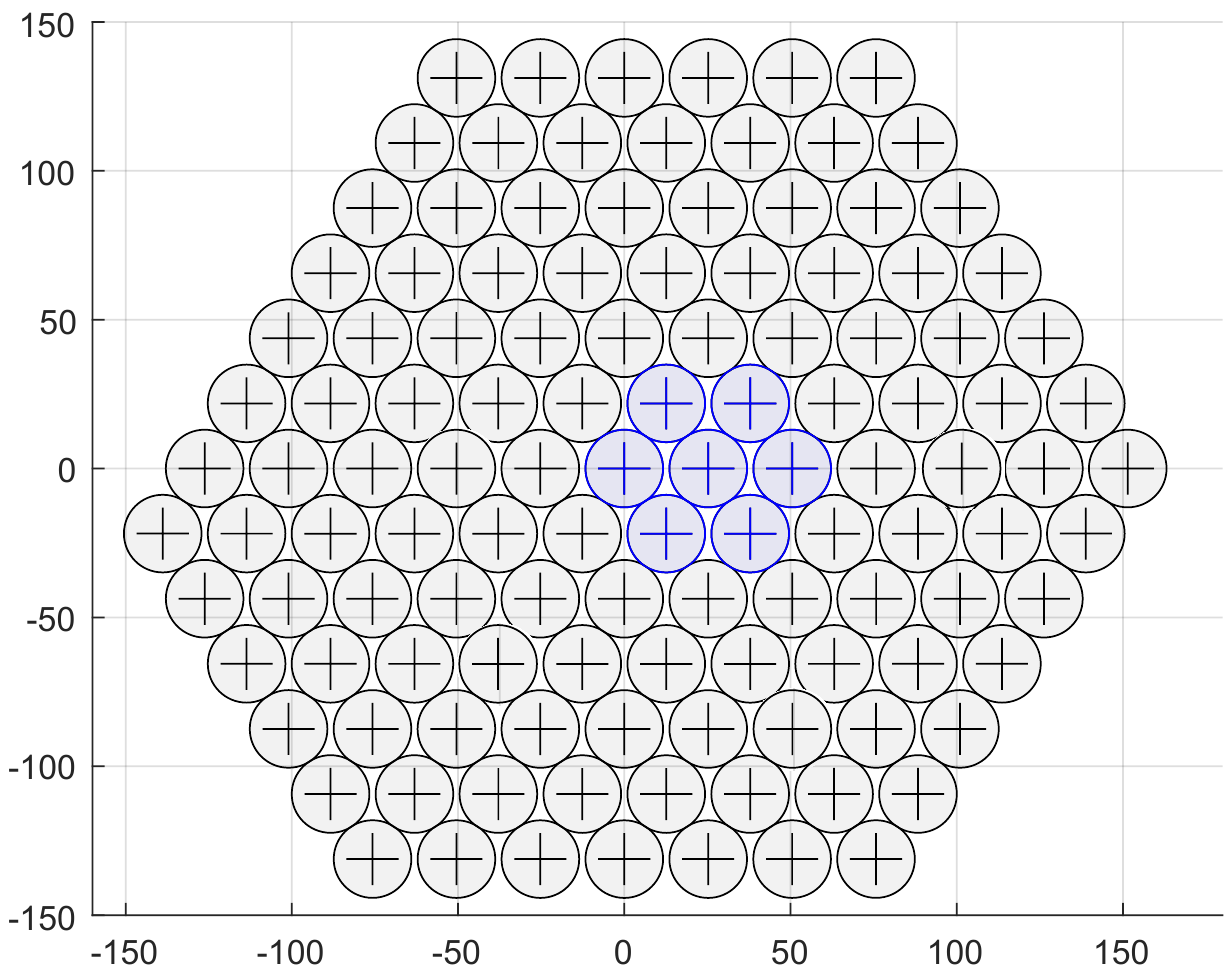}}
		\vspace{-6mm}	
	\end{subfigure}
	\quad
	\begin{subfigure}[b]{.35\textwidth}
		\hspace*{5mm}\subcaptionbox{The localized probability computations within in a typical neighborhood\label{fig:loc}}%
		{\includegraphics[scale=0.35]{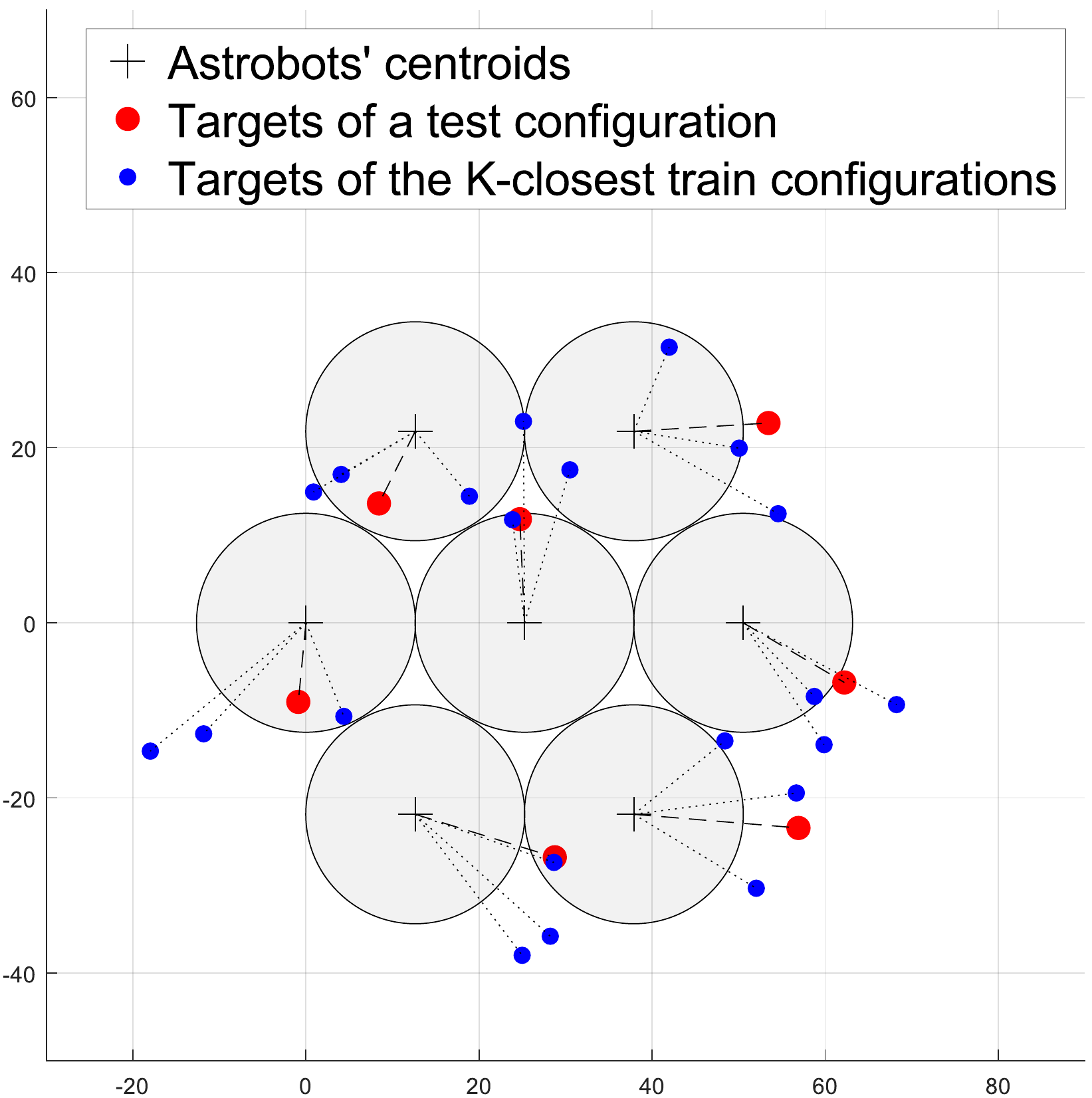}}
	\end{subfigure}
	\caption{A typical probability localization scenario}
\end{figure}

Now, we select the \textit{$k$ closest configurations set}, say, $\bm{P}^{\bm{T},k} \subset \bm{P}$, to $\bm{T}$. The specification of $k$ depends on the size of the train dataset and the complexity of the intended swarm. Namely, it must not be too small, otherwise there is some overfitting risk corresponding to the test configuration. On the other hand, if it is too large, one may take some train configurations into account which do not resemble the desired test one. So, it may lead to inaccurate predictions associated with some astrobots. Assume that function $\text{sort}(\text{set},\text{metric})$ sorts its set argument with respect to its metric argument in ascending order. Moreover, fix function $\text{fetch}(\text{set},k)$ which returns the first $k$ elements of its sorted argument set. Then, given a particular $k$, $\bm{P}^{\bm{T},k}$ is defined as follows.
\begin{equation}
	\bm{P}^{\bm{T},k} \coloneqq \text{fetch}\bigg(\text{sort}\Big(\{\bm{P}_{i} \mid \bm{P}_{i} \in \bm{P} \},\Delta(\bm{T},\bm{P}_{i}) \Big),k\bigg)
\end{equation}
Now, we use weight vector $\bm{w}$, $\bm{P}^{\bm{T},k}$, and its corresponding ground truth vector $\bm{g}^{\bm{P}^{\bm{T},k}} \coloneqq \bigcup\limits_{i}g^{\bm{P}^{\bm{T},k}}_i$ to compute the predictions corresponding to astrobots which converge to configuration $\bm{T}$. One notes that $\bm{w}$ has to be exclusively applied to the 0s in each element of ${}^{\bm{w}}\bm{g}^{\bm{P}^{\bm{T},k}} \coloneqq \bigcup\limits_{i}{}^{\bm{w}}g^{\bm{P}^{\bm{T},k}}_i$. Then, the result is \textit{weighted ground truth vector} whose elements are defined as below.
\begin{equation}
	{}^{\bm{w}}{g}^{\bm{P}^{\bm{T},k}}_{i} \coloneqq 
	\begin{dcases*}
	1 & if ${g}^{\bm{P}^{\bm{T},k}}_i = 1$\\
	{w}_{i} & otherwise
	\end{dcases*}
\end{equation}
Thus, \textit{primary prediction probability vector} $\bm{\hat{\Gamma}}^{\bm{P,T}}$ with respect to test configuration $\bm{T}$ is given by\footnote{Binary operator ($\cdot$)$\oslash$($\cdot$) symbolizes Hadamard division \cite{cyganek2013object}.}
\begin{equation}
	\bm{\hat{\Gamma}}^{\bm{P,T}} \coloneqq \bigg(\sum_{i=1}^{k} \bm{g}^{\bm{P}^{\bm{T},k}}_i\bigg)\oslash\bigg(\sum_{i=1}^{k} {}^{\bm{w}}\bm{g}^{\bm{P}^{\bm{T},k}}_{i}\bigg).
\end{equation}
One may alternatively plan to apply different weights to each ground truth vector with respect to its distance metric from a particular test configuration. However, it increases the risk of overfitting.

If one deals with very large astrobots swarms, the distance metric $\Delta$ may not be reliable to assess the similarity between two configurations. In fact, once the number of astrobots extremely increases, there may be some astrobots among the closest train configuration whose targets are too far from their corresponding ones in the test configuration. This may be problematic even in the case of small swarms. In the next section, we mitigate this issue by localizing the derived prediction probability vector.
\subsection{Prediction Probability Localization}
The global neighborhood analysis of a large astrobots swarm is both inefficient and even problematic in view of the final results. In particular, large swarms geometrically encompass a massive number of neighborhoods. If one checks all available neighborhoods in the course of each lazy evaluation of the algorithm, then the solution would never be obtained after a reasonable amount of time. On the other hand, not all astrobots neighborhoods influence the coordination of a particular astrobot, but only those which are its immediate neighbors. Thus, we have to localize the probability computations of the algorithm. In particular, we perform a local analysis on the neighborhoods of each astrobot. Thereby, the algorithm is exclusively applied to a number of small configurations which includes a maximum number of 7 astrobots. By doing so, it would be less likely to have some astrobots the distances between whose targets and a test configuration are high. For example, Fig. \ref{fig:neigh} depicts a neighborhood of astrobots the magnitudes of whose metric distances are reasonable as illustrated in Fig. \ref{fig:loc}.

Let $\bm{P}$ be a configuration of including $n$ astrobots. We define neighborhood $\bm{\nu_{\pi}} \subset \bm{P}$ associated with a typical astrobot $\pi$ as the central entity of this neighborhood. The dimension of each instance of $\bm{\nu_{\pi}}$ is $2\times r$, where $1\le r \le 7$ denotes the number of the astrobots in the neighborhood. Thus, one has to overall perform $n$ local analyses. To do so, we introduce \textit{counter vector} $\bm{\eta}$ whose dimension is $1\times n$. Element ${\eta}_{i}$ of $\bm{\eta}$ corresponds to the number of the neighborhoods to which the $i$th astrobot of the swarm belongs. The elements of $\bm{\eta}$ are integers varying between 1 and 7. Thus, we yield \textit{neighborhood probability vector} $\widetilde{\bm{\Gamma}}^{\bm{\nu_{\pi},T}}$ with respect to neighborhood $\bm{\nu_{\pi}}$ whose elements are defined as follows.
\begin{equation}
\widetilde{\Gamma}^{\bm{\nu_{\pi},T}}_{i} \coloneqq 
\begin{dcases*}
{\hat{\Gamma}}^{\bm{\nu_{\pi},T}}_{i} & if ${\pi}_{i} \in \bm{\nu_\pi}$\\
0 & otherwise
\end{dcases*}
\end{equation}
Now, given, $\widetilde{\bm{\Gamma}}^{\bm{\nu_{\pi},T}}\coloneqq \bigcup\limits_{i} \widetilde{\Gamma}^{\bm{\nu_{\pi},T}}_i$, \textit{final probability vector} $\bm{\Gamma^{P,T}}$ is computed as
\begin{equation}
	\bm{\Gamma^{P,T}}\coloneqq \left[\sum\limits_{\bm{\pi}}\widetilde{\bm{\Gamma}}^{\bm{\nu_{\pi},T}}\right]  \oslash \bm{\eta}
\end{equation}
 Finally, we need to transform these probabilistic entries to categorical 1s and 0s to represent successful or failed convergences, respectively. For this purpose, given a \textit{decision filter} $q$, we define the elements of \textit{output vector} $\bm{Y} \coloneqq \bigcup\limits_{i} y_{i}$ as below.
\begin{equation}
y_{i} \coloneqq 
\begin{dcases*}
1 & if ${\Gamma}_i^{\bm{P,T}} > q$\\
0 & otherwise
\end{dcases*}
\end{equation}

The last phase of our convergence prediction algorithm performs Monte Carlo cross-validation to assess the credibility of the results. The rational behind preferring this method over $k$-fold cross-validation is the computational efficiency of the former. Namely, Monte Carlo cross validation enjoys a property that the proportional relation between train/test splits does not depend on the number of validation iterations. Thus, one can perform a series of iterations which are not linked to the dimensions of train and test datasets. The drawback of this method, though, is that some configurations may never be selected as test configurations, whereas others may be selected multiple times. For this reason, it is necessary to put a particular attention to the number of validation iterations. The choice of this number depends on how large a typical test dataset is compared to its corresponding train one. In other words, the smaller the test dataset is, the larger the number of iterations has to be.
\section{Simulations and Results}
\label{sec:sim}
In this section\footnote{The simulation scripts were all written in Matlab\texttrademark 2019 and performed on a Dell Inspiron 15 7000 which is supported by an Intel Core i7-7700HQ processor with 2.80 GHz clockspeed, 16GB RAM, and Windows 10 Home 64 bit.}, we illustrate how our algorithm efficiently predict the convergence of astrobots in massive swarms.  We first define our evaluation measures and performance metrics. Then, we take two swarms into account each of which is constituted by 116 and 487 astrobots. The latter is particularly similar to the case of the astrobots swarm associated with the SDSS-V project \cite{kollmeier2017sdss}. We also present some hints regarding the value selections for the algorithm's hyperparameters.
\subsection{Performance Measures}
Our performance measures are essentially defined based the following four notions.
\begin{itemize}
	\item A true positive (TP) is an astrobot which is predicted to converge (the predictor predicts 1), and it actually converges to its target position (its corresponding ground truth element is 1).
	\item A false positive (FP) is an astrobot which is predicted to converge (the predictor predicts 1), but it actually does not converge to its target position (its corresponding ground truth element is 0).
	\item A true negative (TN) is an astrobot which is not predicted to converge (the predictor predicts 0), and it actually does not converge to its target position (its corresponding ground truth element is 0)
	\item A false negative (FN) is an astrobot which is not predicted to converge (the predictor predicts 0), but it actually converges to its target position (its corresponding ground truth element is 1).
\end{itemize} 
\begin{figure}
	\centering
	\hspace*{-1cm}\includegraphics[scale=0.6]{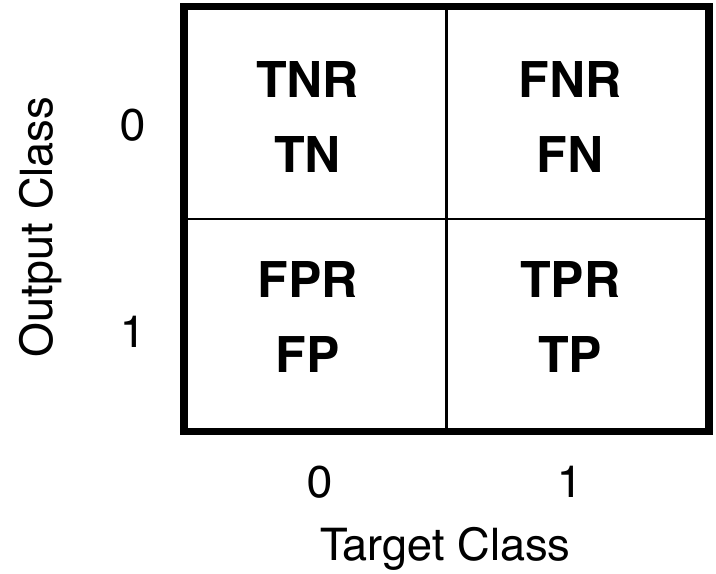}
	\caption{The layout of the information represented by a typical confusion matrix}
	\label{fig:conf}
\end{figure}
Accordingly, we take the standard rates of the above factors, i.e., TPR, FPR, TNR, and FNR, into account. These values are reported in confusion matrices based on Fig. \ref{fig:conf}. If a predictor yields good performance, its corresponding confusion matrix has large values in its main-diagonal entries, indicating that the majority of samples have been correctly classified.
\begin{figure*}
	\centering
	\hspace*{0mm}
	\begin{subfigure}[b]{.24\textwidth}
		\hspace*{-0mm}\subcaptionbox{$k = 3$\label{fig:166-3}}
		{\includegraphics[scale=0.3]{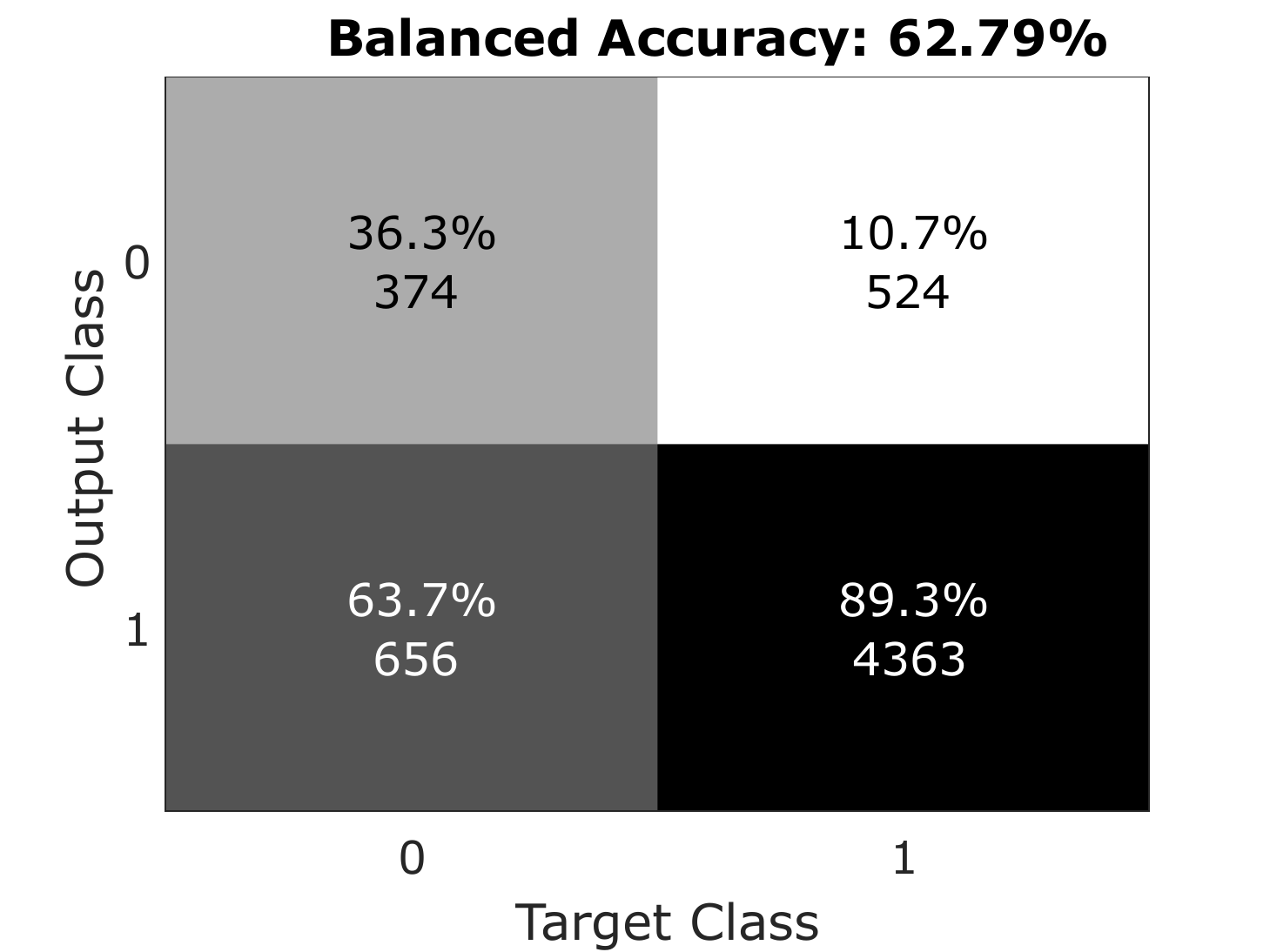}}
		\vspace{-5.5mm}	
	\end{subfigure}%
	\begin{subfigure}[b]{.24\textwidth}
		\hspace*{0mm}\subcaptionbox{$k = 13$\label{fig:166-13}}
		{\includegraphics[scale=0.3]{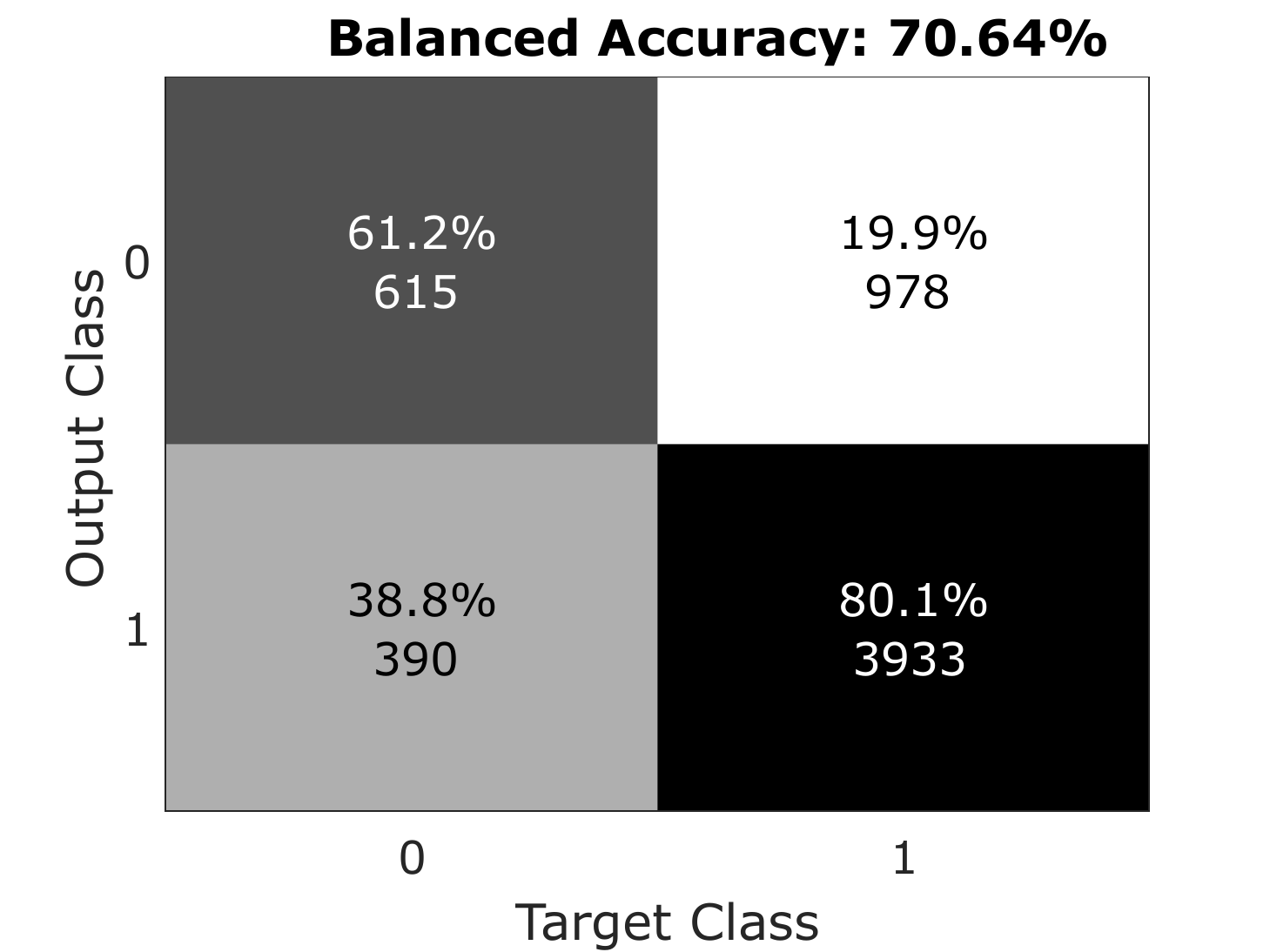}}
	\end{subfigure}%
	\begin{subfigure}[b]{.24\textwidth}
		\hspace*{0mm}\subcaptionbox{$k = 25$\label{fig:166-25}}
		{\includegraphics[scale=0.3]{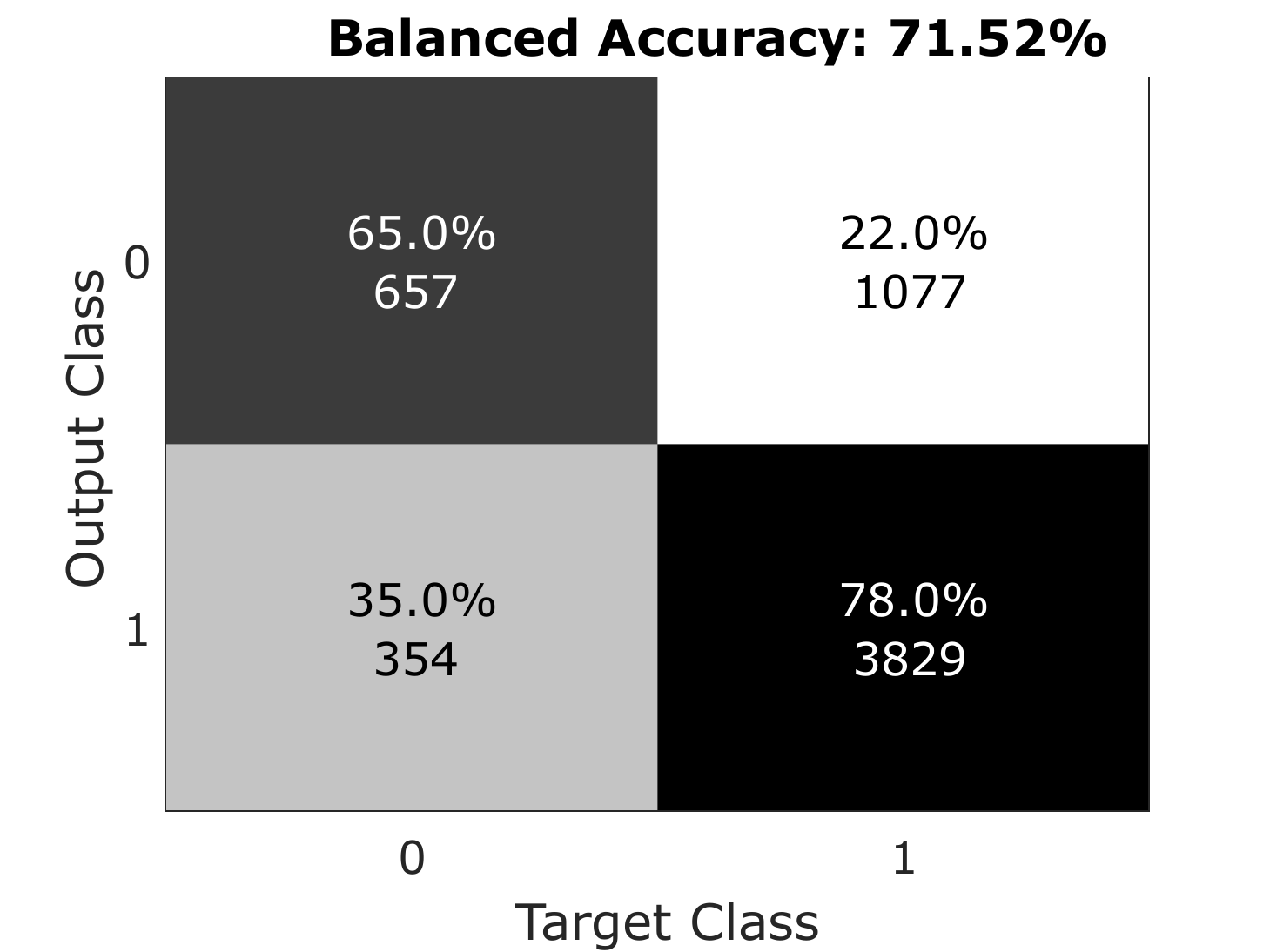}}
	\end{subfigure}%
	\begin{subfigure}[b]{.24\textwidth}
		\hspace*{0mm}\subcaptionbox{$k = 39$\label{fig:166-39}}
		{\includegraphics[scale=0.3]{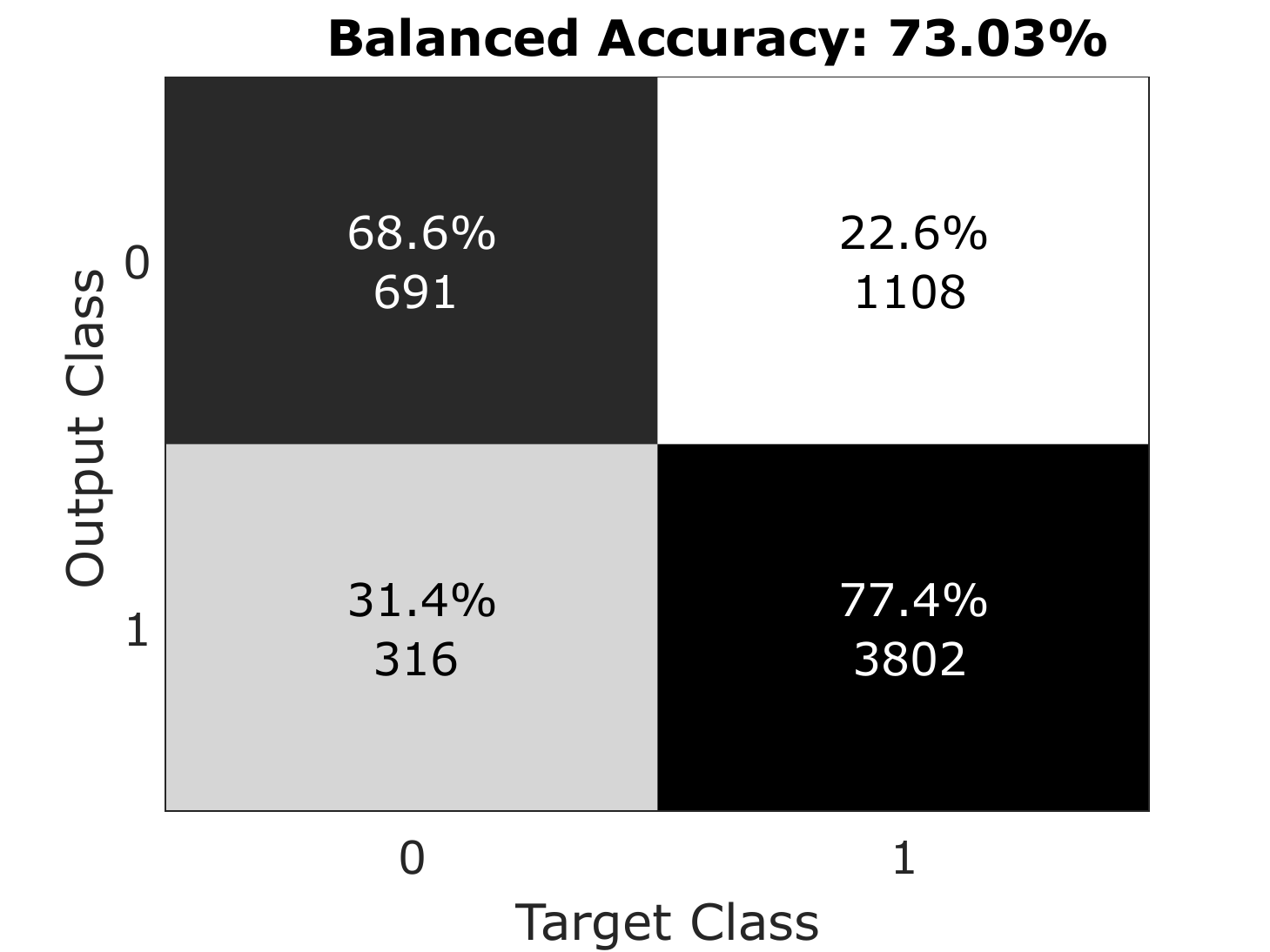}}
	\end{subfigure}
	\caption{The confusion matrices of the 116-astrobots swarm}
	\label{fig:116-confmat}
\end{figure*}
\begin{figure*}
	\centering
	\hspace*{-15mm}
	\begin{subfigure}[]{.48\textwidth}
		\hspace*{-0mm}\subcaptionbox{$k = 13$\label{fig:166-av-13}}
		{\includegraphics[scale=0.24]{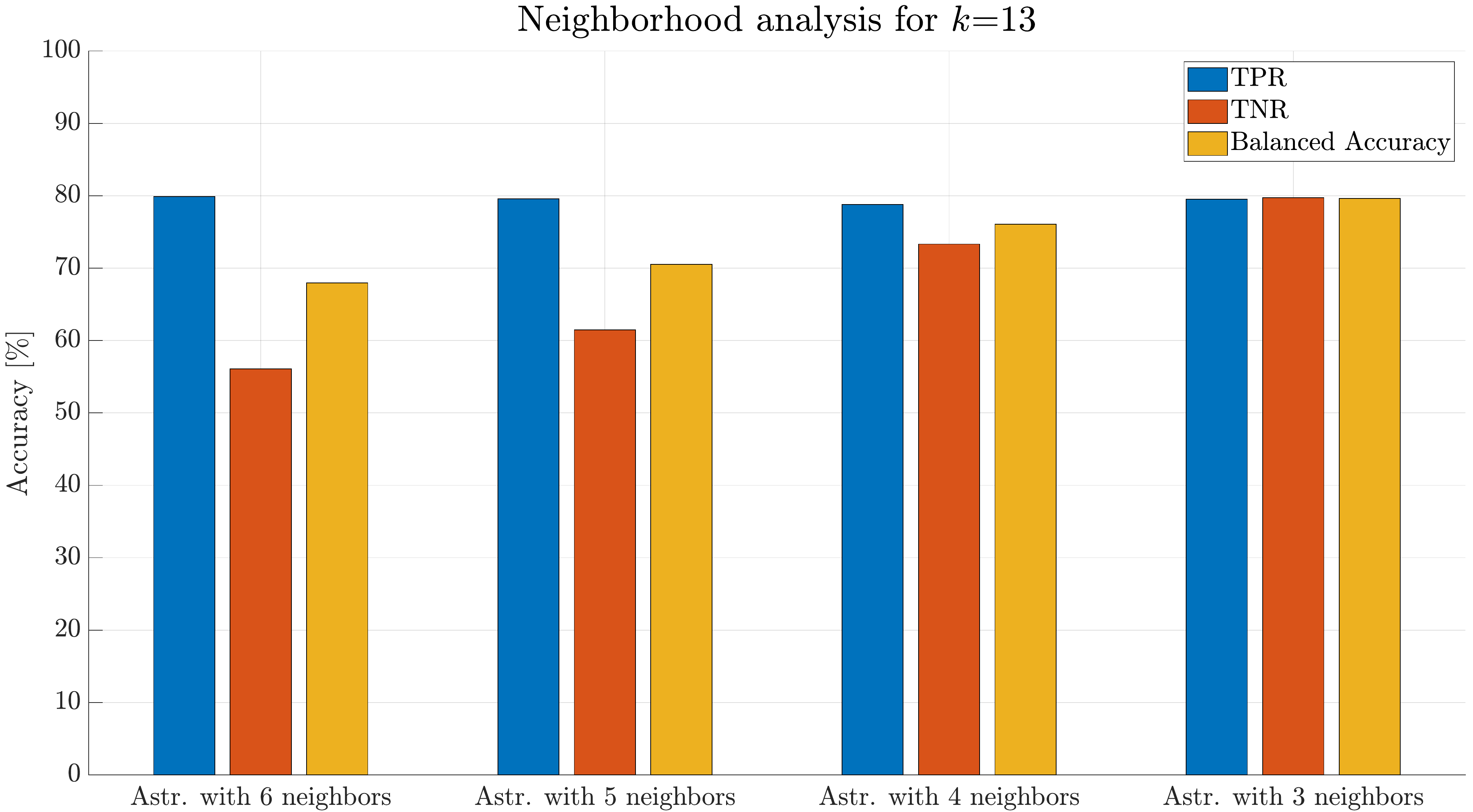}}
		\vspace{-0mm}	
	\end{subfigure}\quad
	\begin{subfigure}[]{.48\textwidth}
		\hspace*{5mm}\subcaptionbox{$k = 39$\label{fig:166-av-39}}
		{\includegraphics[scale=0.24]{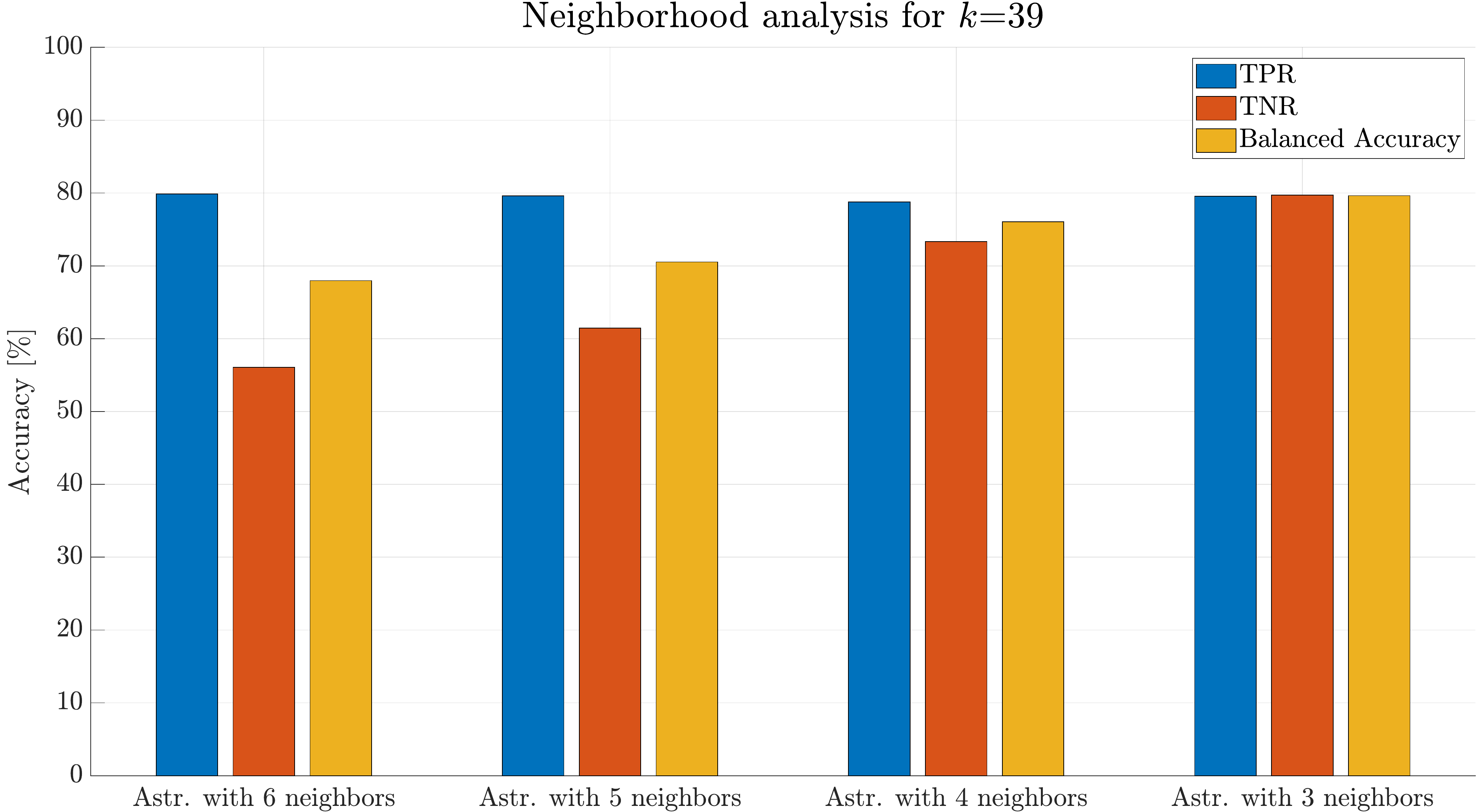}}
	\end{subfigure}
	\caption{The neighborhood analyses of the 116-astrobots swarm corresponding to two neighborhood selections}
	\label{fig:116-neigh-ana}
\end{figure*}

From an engineering point of view, we are more interested in the correct predictions of positives (the astrobots which converge to their target positions). It is because the information regarding the prediction of these astrobots would be crucial to decide whether or not a coordination process should be executed associated with a particular configuration of targets. On the other hand,  the number of positives is much greater than that of negatives. If the predictor always predicts 1, the TPR would be perfect. But, the predictor does not indeed predict anything by completely neglecting 0s. So, we track the balanced accuracy measure established as the average of the TPR and the TNR to better assess the predictive essence of the algorithm. We also employ receiver operating characteristic (ROC) curves to illustrate the performance of our predictor in the course of varying one of its hyperparameters. 

We also take precision and F1 score (harmonic mean) into the consideration. The precision measure is an index of how accurate the predictor is in predicting positives. Precision is an important measure to look at when FPs have significant impacts on our problem. We intend to maximize precision through minimization of FPs. F1 score indicates the trade-off between precision and the TPR, say, recall. For example, if we increase the TPR, we indeed increase the number of predicted TPs. However, we also increase the number of FPs, thereby decreasing the precision. The bigger the F1 score is, the better the trade-off between precision and recall is.

We include corrector coefficients $\alpha$ and $\beta$, as well. These hyperparameters are used to manually tune the weight vector $\bm{w}$ to obtain better accuracy rates with respect to positives and negatives. In particular, $\alpha$ tunes the ${w}_{i}$s of the astrobots in total neighborhoods, while $\beta$ does the same but for the astrobots residing in partial neighborhoods. In all simulations, we take the decision filter $q = 0.5$.
\subsection{A swarm including 116 astrobots}
\begin{figure*}
	\centering
	\hspace*{0mm}
	\begin{subfigure}[]{.23\textwidth}
		\hspace*{-0mm}\subcaptionbox{$\alpha = \beta = 0.95$\label{fig:166-ab-095}}
		{\includegraphics[scale=0.3]{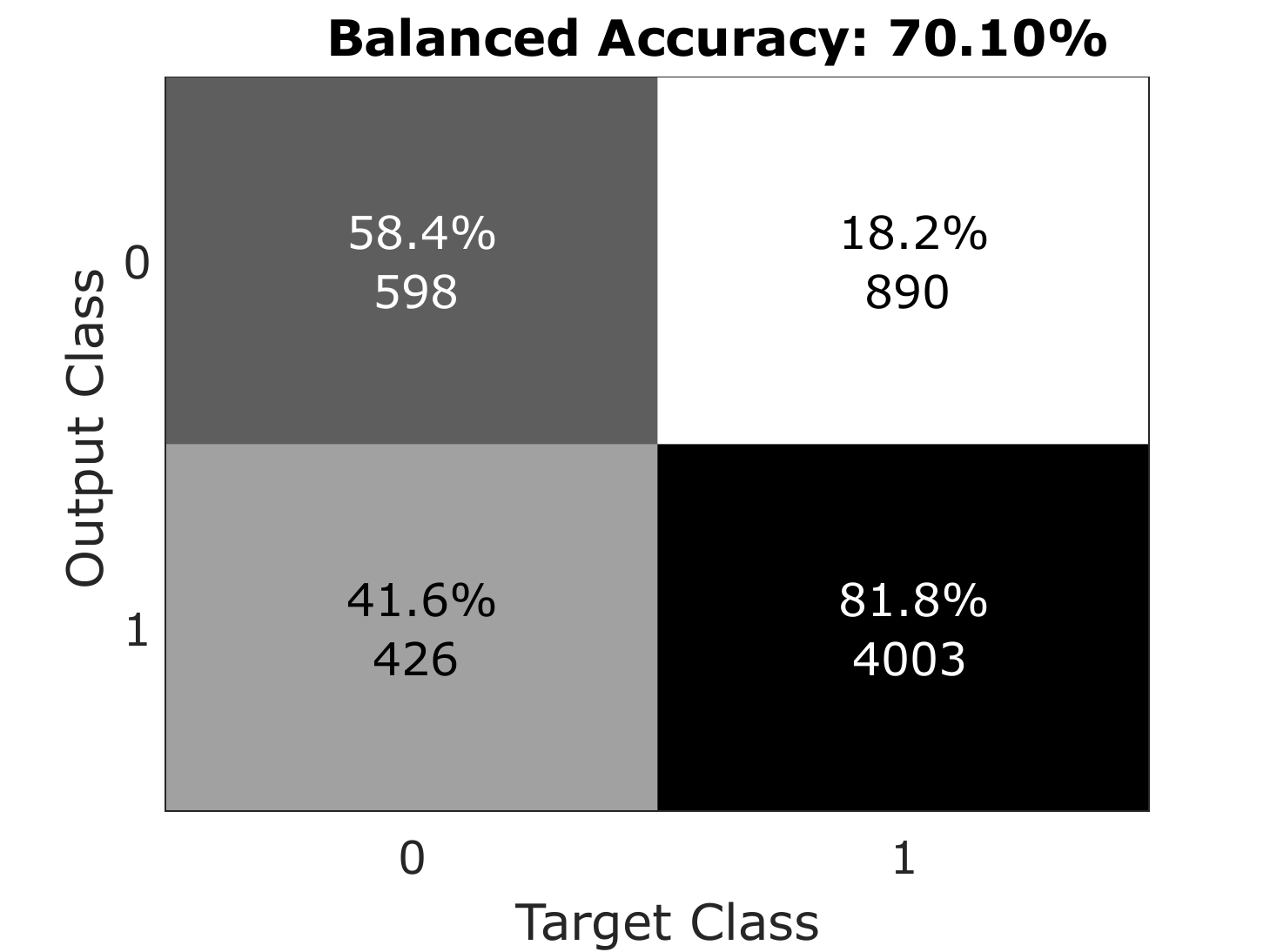}}
		\vspace{-0mm}	
	\end{subfigure}%
	\begin{subfigure}[]{.23\textwidth}
		\hspace*{0mm}\subcaptionbox{$\alpha = \beta = 1.00$\label{fig:166-ab-100}}
		{\includegraphics[scale=0.3]{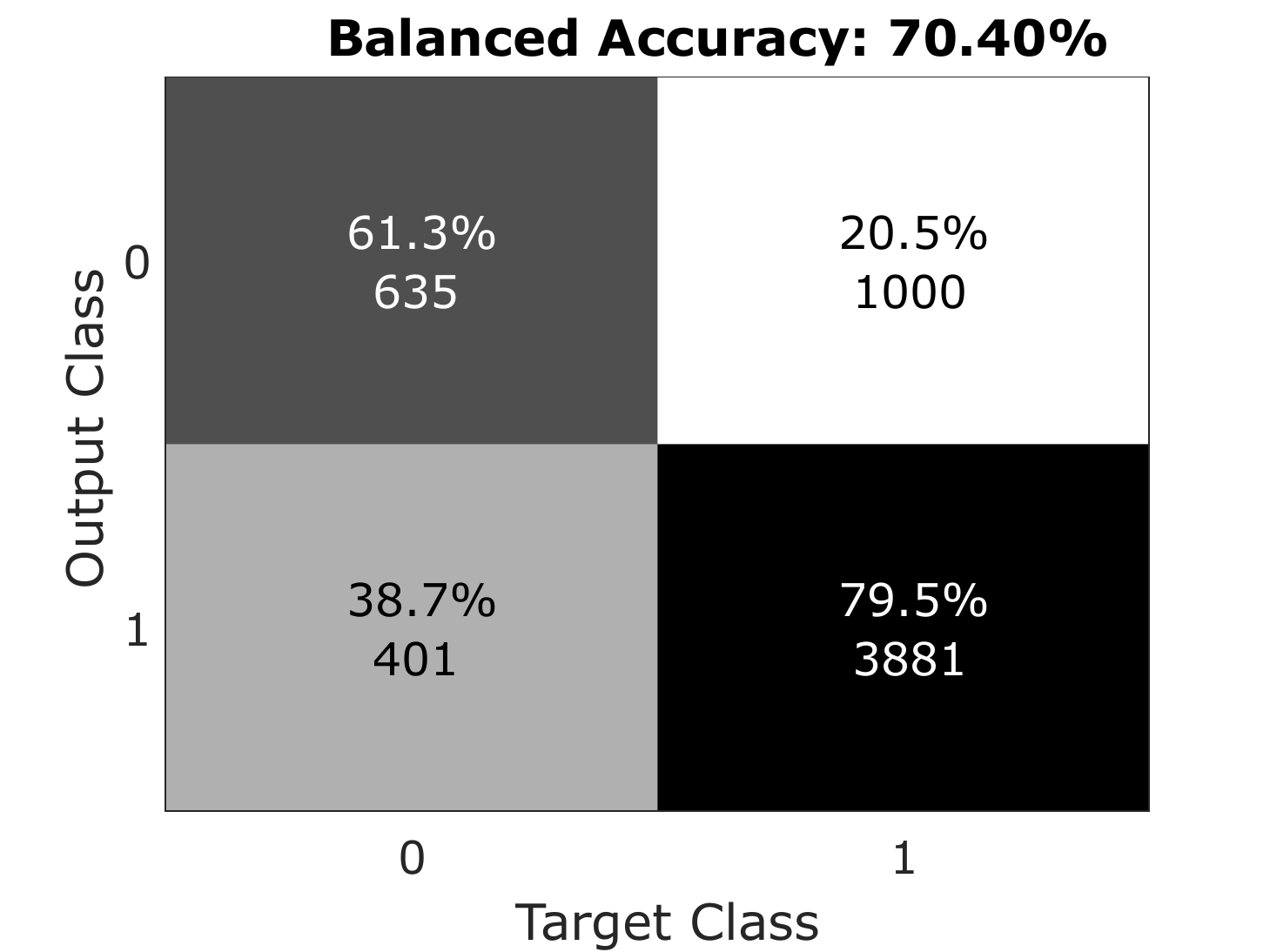}}
	\end{subfigure}%
	\begin{subfigure}[]{.23\textwidth}
		\hspace*{0mm}\subcaptionbox{$\alpha = \beta = 1.05$\label{fig:166-ab-105}}
		{\includegraphics[scale=0.3]{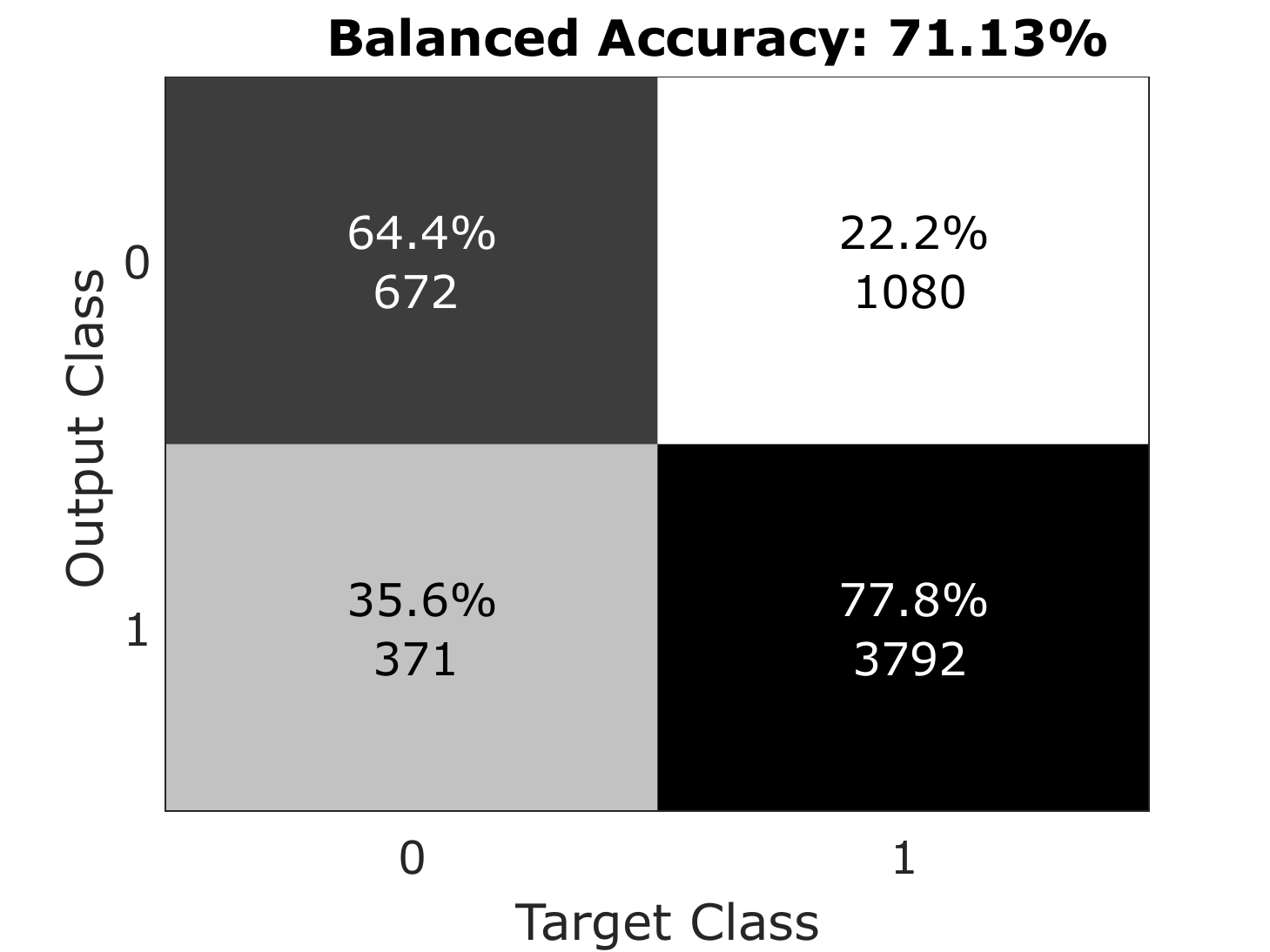}}
	\end{subfigure}
	\caption{The evolution of the confusion matrices of the 116-astrobots swarm given different corrector factors}
	\label{fig:116-ab}
\end{figure*}
\begin{figure*}
	\centering
	\hspace*{-15mm}
	\begin{subfigure}[]{.33\textwidth}
		\hspace*{-0mm}\subcaptionbox{$\alpha= \beta = 1$\label{fig:166-k-ana}}
		{\includegraphics[scale=0.25]{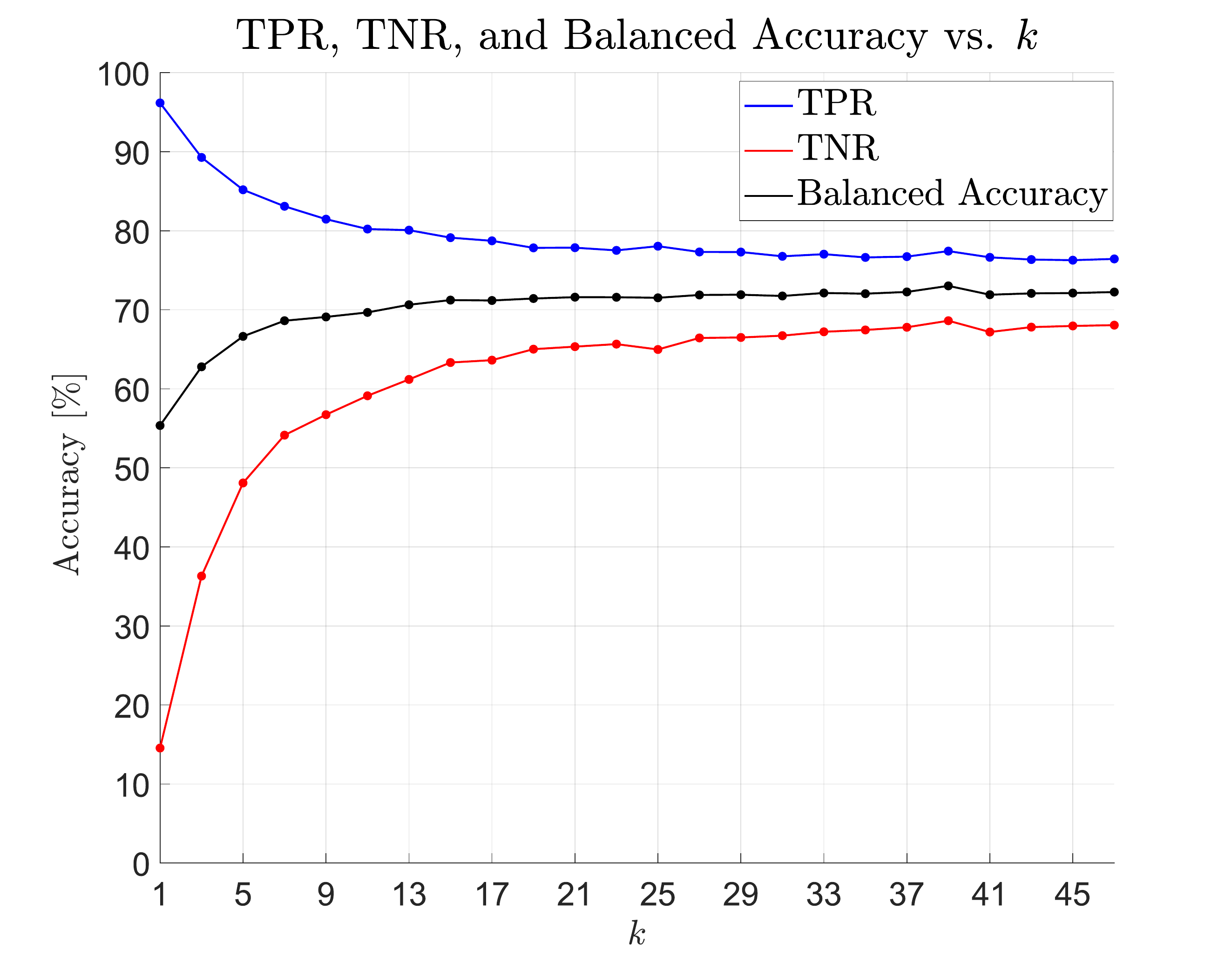}}
		\vspace{-0mm}	
	\end{subfigure}%
	\begin{subfigure}[]{.33\textwidth}
		\hspace*{5mm}\subcaptionbox{$k = 13$\label{fig:166-w-ana}}
		{\includegraphics[scale=0.25]{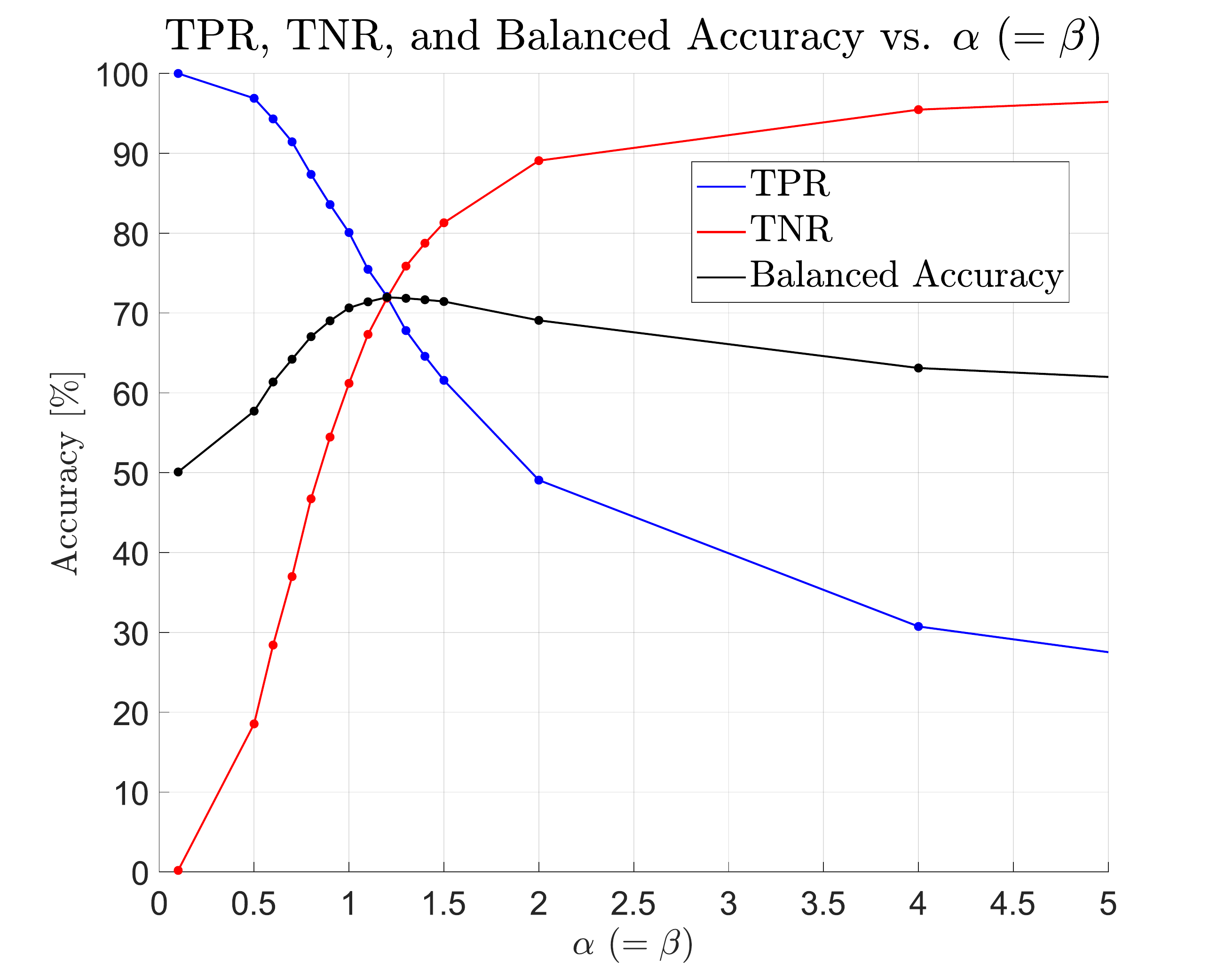}}
	\end{subfigure}%
	\begin{subfigure}[]{.33\textwidth}
		\hspace*{5mm}\subcaptionbox{$\alpha= \beta = 1$\label{fig:166-prf1}}
		{\includegraphics[scale=0.25]{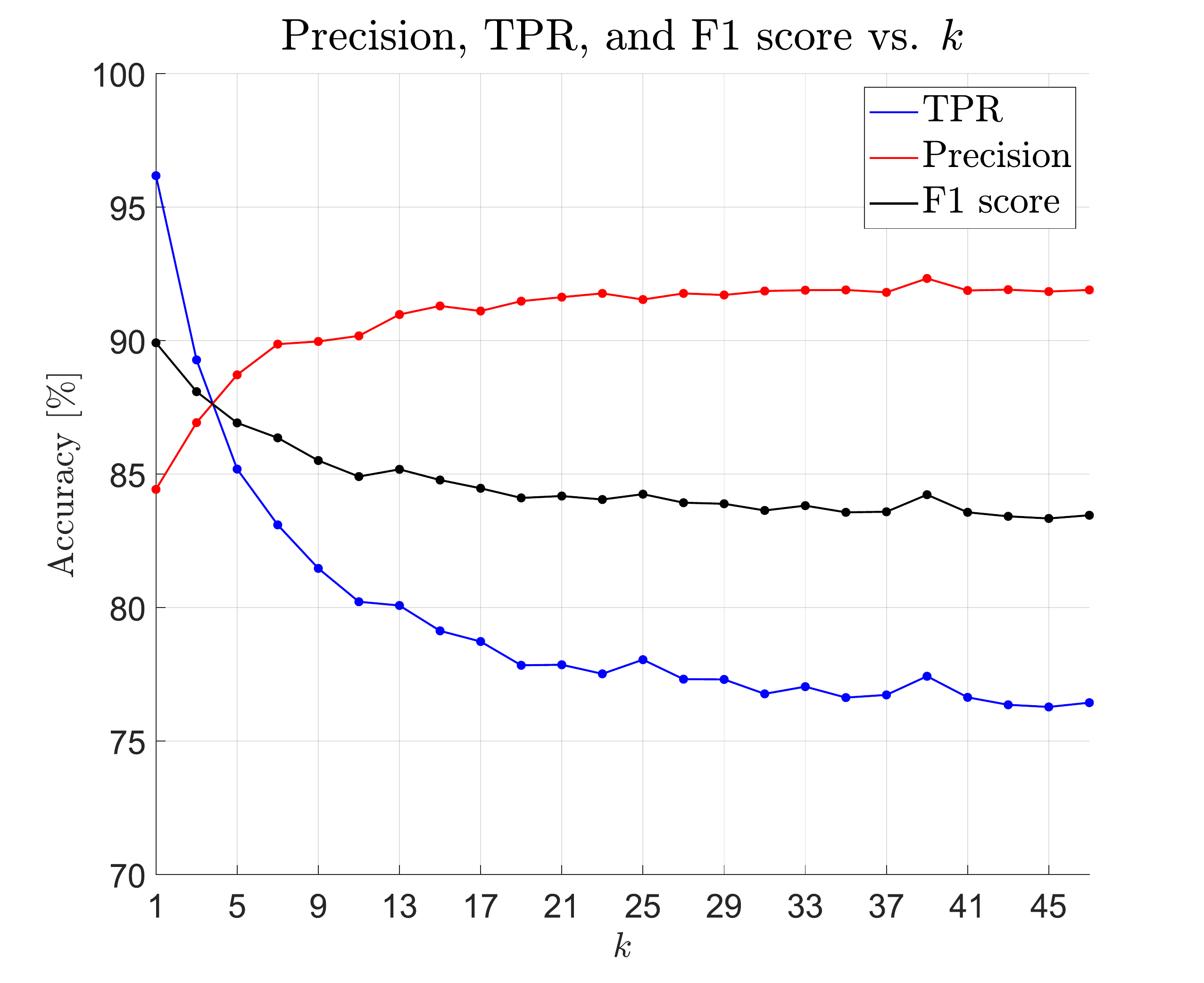}}
	\end{subfigure}
	\caption{Accuracy measures for the 116-astrobots swarm}
	\label{fig:116-ref}
\end{figure*}
Our complete dataset is composed of 10100 configurations, where the train and the test datasets include 10049 and 51 configurations, respectively. The algorithm iterations is set to 15. The confusion matrices corresponding to various values of $k$ are depicted in Fig. \ref{fig:116-confmat}. We observe that increasing $k$ increases and decreases the TNR and the TPR, respectively. It is reasonable since the more train configurations we take into account for the computation of the output, the higher the likelihood is to consider the train configurations whose astrobots don’t converge. The selection of $k$ depends on how large the train dataset is. The larger the train dataset is, the larger $k$ may be. In this scenario, a proper $k$ may be chosen in the range of 10 to 50. If we increase $k$ too much, the information about the targets locations of the closest train configurations are no longer reliable.

It is also interesting to assess our performance indices for single astrobots. In particular, we obtain the TPR, the TNR, and the balanced accuracy, on the basis of the number of each astrobot's neighbors. To do so, we take the average of the performances of the astrobots with a specific number of neighbors, as rendered in Fig. \ref{fig:116-neigh-ana}, where both corrector coefficients are 1. Namely, Fig. \ref{fig:166-av-13} indicates that the prediction accuracy bottleneck refers to the astrobots in total neighborhoods. Fig. \ref{fig:166-av-39} illustrates how the the balanced accuracy is improved in total neighbourhoods. On the other hand, the astrobots of partial neighborhoods experience the decrement and the increment of the TPR and the TNR, respectively.

Corrector coefficients are expected to impact the qualities of the cases in which total neighborhoods are fairly abundant. In Fig. \ref{fig:116-ab}, the confusion matrices of three different predictions are reported in which corrector coefficients are varied. In this case, we simply keep their values the same to show the overall effect of magnifying the weights of 0s. In all of the cases, we have $k=13$. In particular, it is evident that increasing the corrector coefficients leads to the increment and the decrement of TNR and TPR, respectively, which is a direct effect of increasing the weight of the minority class.

\begin{figure}[!b]
	\centering
	\includegraphics[scale=0.3]{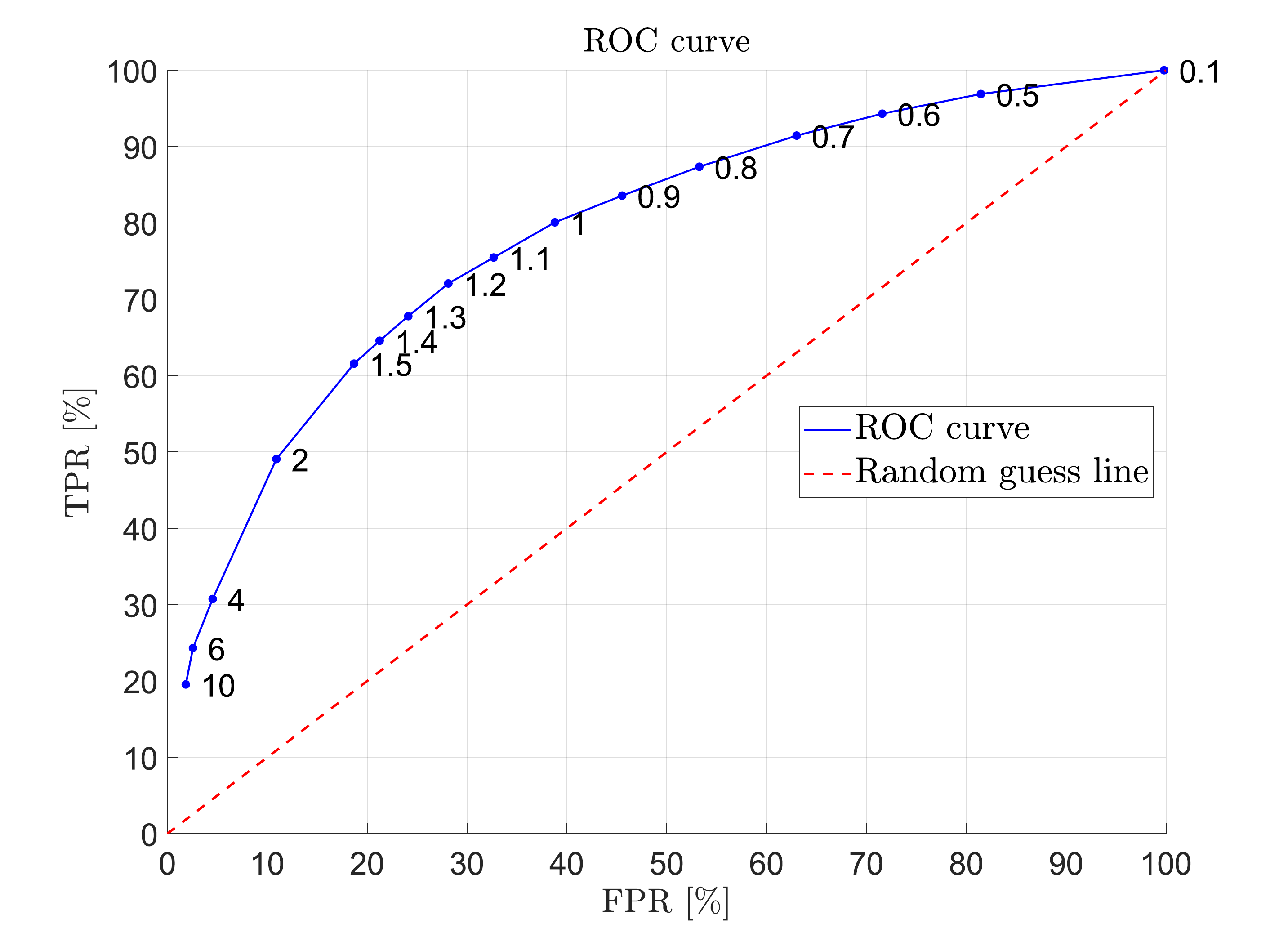}
	\caption{The ROC curve associated with the 116-astrobots swarm}
	\label{fig:116-roc}
\end{figure}

To tune the hyperparameters, one may find Fig.~\ref{fig:116-ref} very useful. Fig.~\ref{fig:166-k-ana} illustrates that any $k>21$ is stable. Specially, $k=39$ realizes the best predictive performance for this swarm. One may note that the right choice of $k$ directly depends on what factor is the main goal of the prediction to be improved. For example, if we one would like to increase the balanced accuracy as much as possible, yet allow the TPR to drop under $80\%$, then $k=39$ seems to be the best choice. But, if the TPR has to be over $80\%$ with the maximum balanced accuracy, one may pick $k=13$. The dynamical trends of the TPR, the TNR, and balanced accuracy are also evident in Fig. \ref{fig:166-w-ana} in the course varying the two corrector coefficients while fixing $k=13$. So, since we are more interested in the correct predictions of the positives, $k$ may be chosen large to increase the TPR as much as possible, while assuring that the balanced accuracy does not drop below a certain threshold. Moreover, Fig.~\ref{fig:166-prf1} shows the trends of the precision, the recall and the F1 score for different values of $k$. Table \ref{tbl:1} reflects the best results in the convergence predictions of the 116-astrobots swarm. Finally, we can look at the ROC curve which visualizes the performance of our predictor. Every point on the ROC curve represents the result of a prediction experiment using a different value of $\alpha(=\beta)$ as shown in Fig.~\ref{fig:116-roc}. Here, we have $k=13$. 
\subsection{A swarm including 487 astrobots}
The qualities of the results in this case fairly follows the qualities of the 116-astrobots case. So, we observe that our algorithm performance remains relatively acceptable even in the case of very complex swarms. To support this claim, we consider a dataset including 10100 configurations, 10049 of which are used to train the predictor and the remaining 51 ones are test configurations. The number of iterations are 15.
\begin{table*}
	\centering
	\caption{The best prediction results corresponding to the 116- and 487-astrobots swarms}
	\begin{tabular}{c@{\hskip 0.3in}ccc@{\hskip 0.3in}ccc@{\hskip 0.3in}cc}
		\toprule  
		Swarm population&K &$\alpha$&$\beta$&TPR(\%)&TNR(\%)&Balanced accuracy(\%)&Precision(\%)&F1(\%)\\\cmidrule{1-9}
		\multirow{3}{*}{116}&31&1&0.9&79.3&64.7&72&91.51&84.97\\
		&39&1&0.85&88.44&63.23&71.83&91.4&85.57\\
		&39&1&1&77.2&68.51&72.85&92.22&84.04\\\cmidrule{1-9}
		\multirow{3}{*}{487}&39&1&0.9&79.94&60.73&70.33&89.51&84.45\\
		&51&1&0.88&80.20&60.97&70.59&89.52&84.61\\
		&51&1&1&78.23&63&70.62&89.78&83.61\\
		\bottomrule		
	\end{tabular}
	\label{tbl:1}
\end{table*}
\begin{figure*}[]
	\centering
	\hspace*{0mm}
	\begin{subfigure}[]{.25\textwidth}
		\hspace*{-0mm}\subcaptionbox{$\alpha = \beta = 0.95$\label{fig:487-con13}}
		{\includegraphics[scale=0.3]{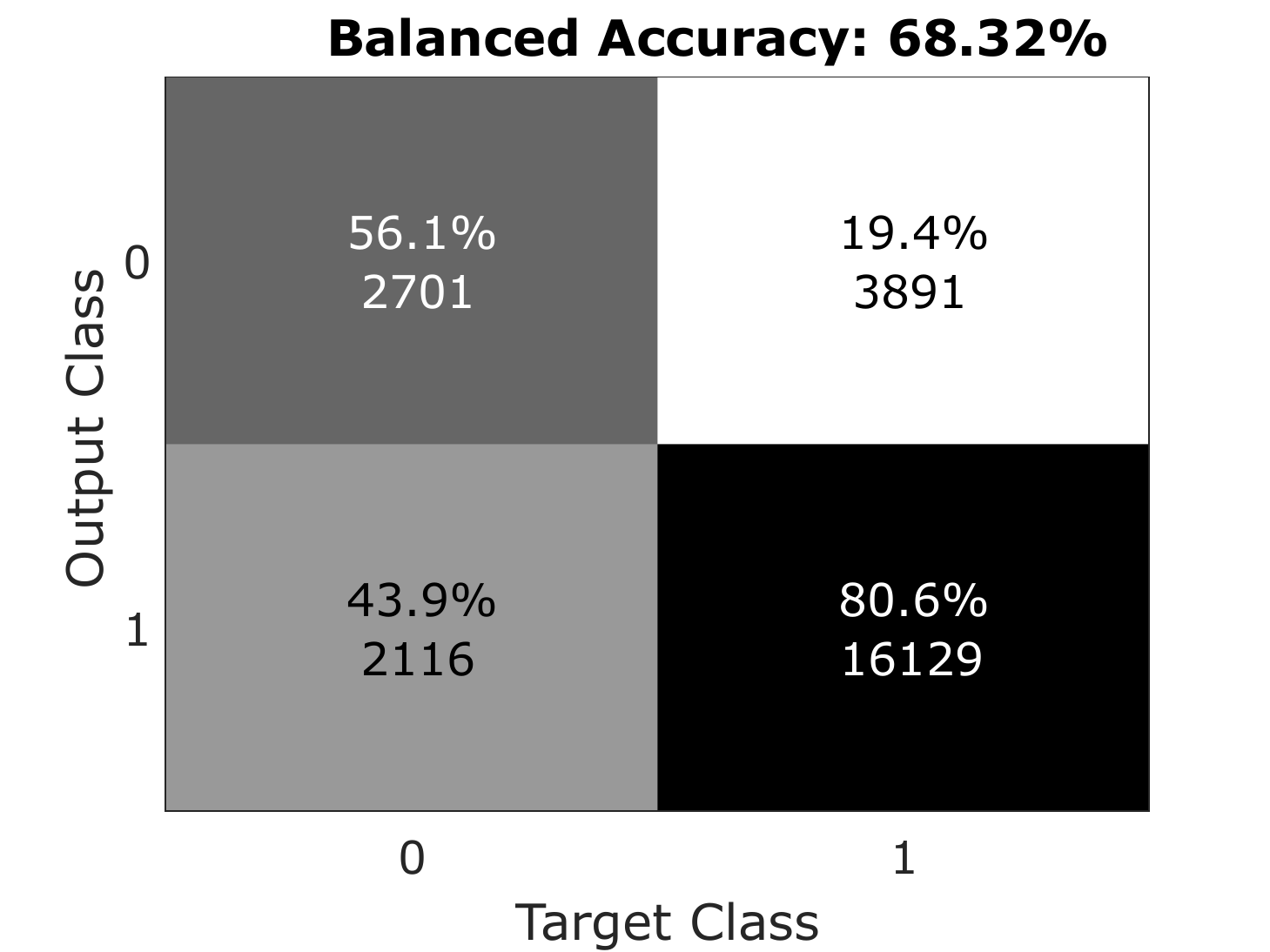}}
		\hspace*{0mm}\subcaptionbox{$\alpha = \beta = 1.00$\label{fig:487-con39}}
		{\includegraphics[scale=0.3]{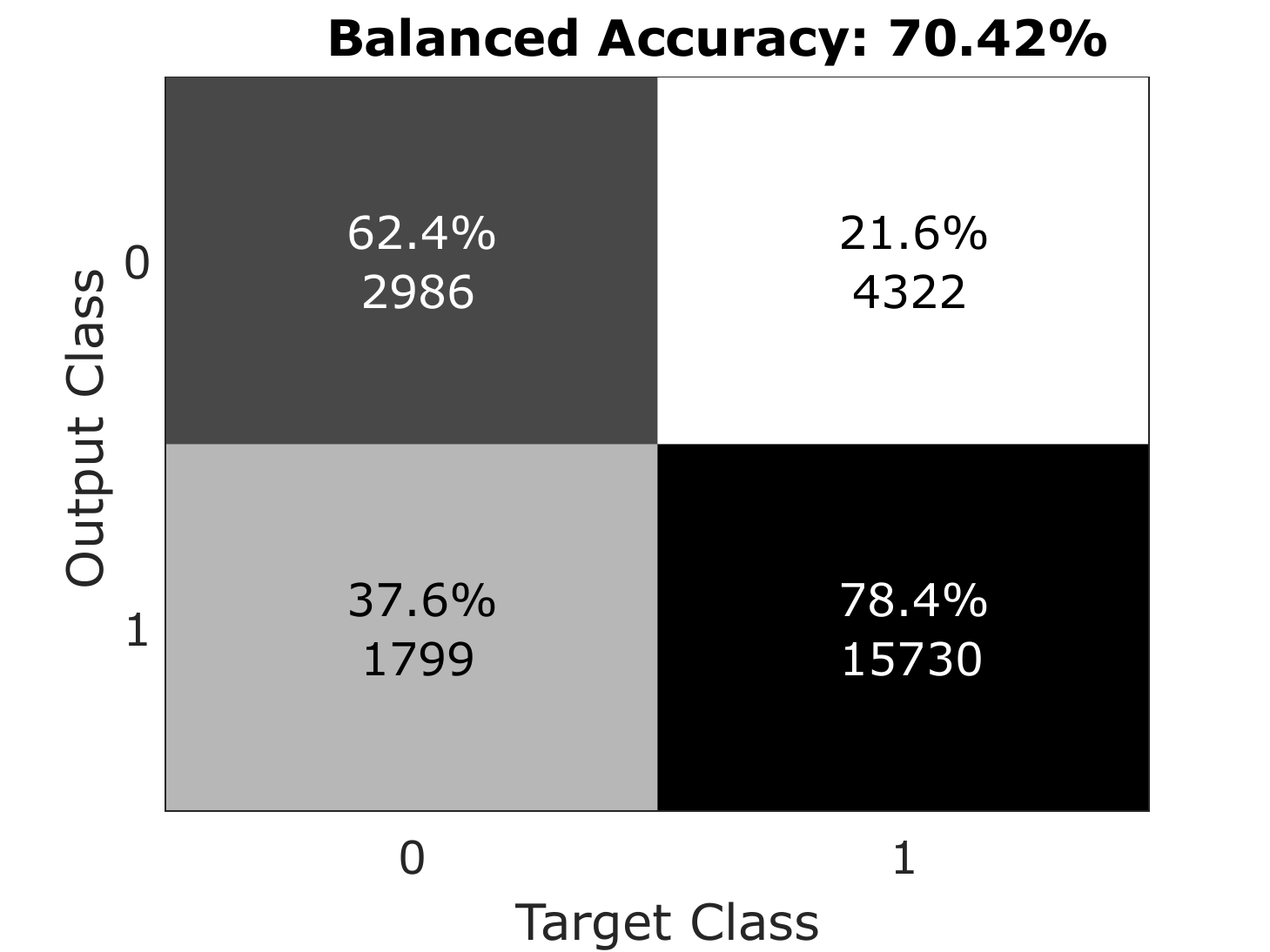}}
		\vspace{-0mm}	
	\end{subfigure}%
	\begin{subfigure}[]{.65\textwidth}
		\hspace*{0mm}\subcaptionbox{$\alpha = \beta = 1.05$\label{fig:487-neigh}}
		{\includegraphics[scale=0.3]{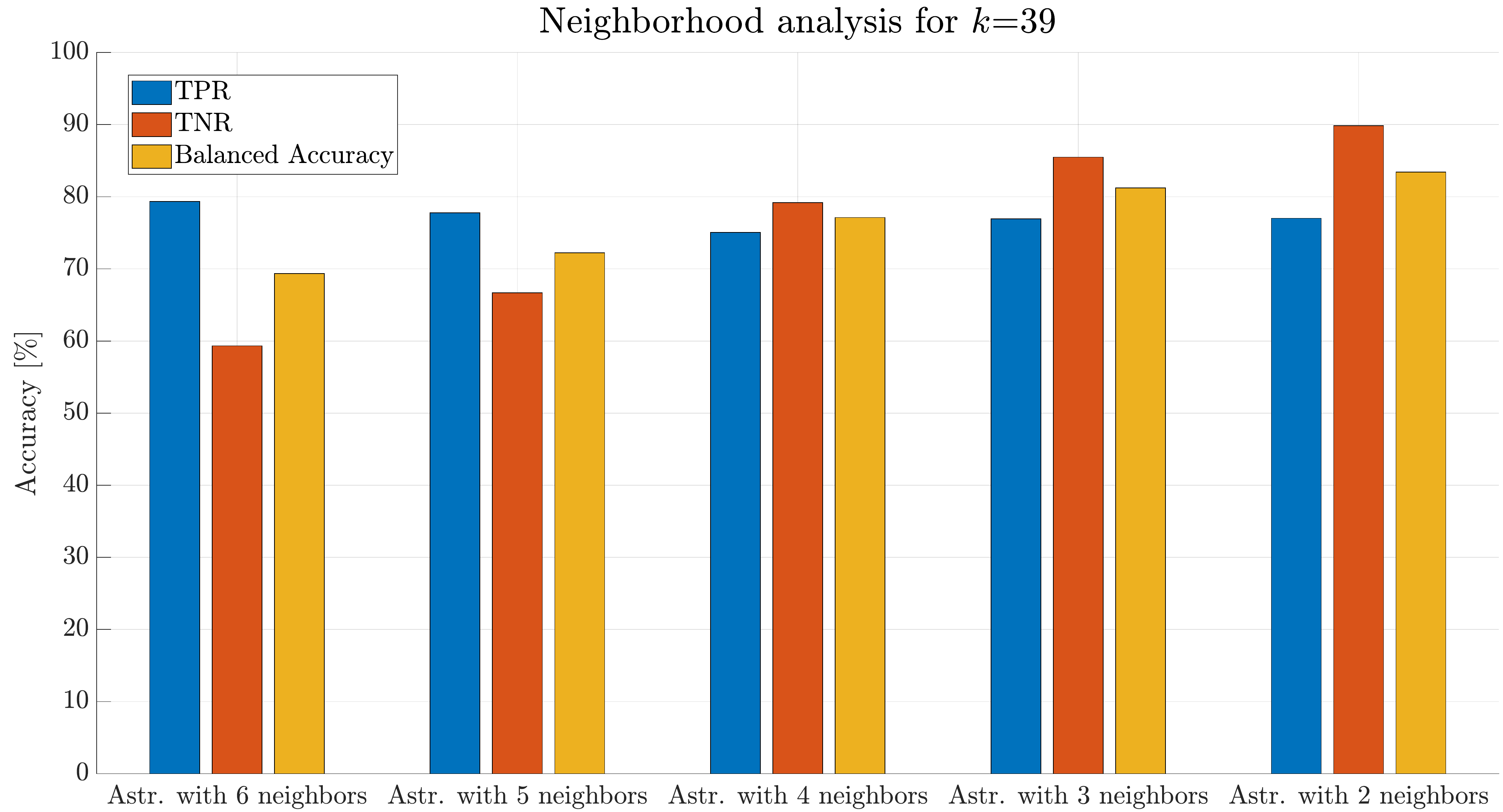}}
	\end{subfigure}
	\caption{Two confusion matrices and a neighborhood analysis corresponding to the 487-astrobots swarm}
	\label{fig:487-con-neigh}
\end{figure*}
\begin{figure*}
	\centering
	\hspace*{0mm}
	\begin{subfigure}[]{.45\textwidth}
		\hspace*{-0mm}\subcaptionbox{$\alpha = \beta = 1$ \label{fig:487-k}}
		{\includegraphics[scale=0.27]{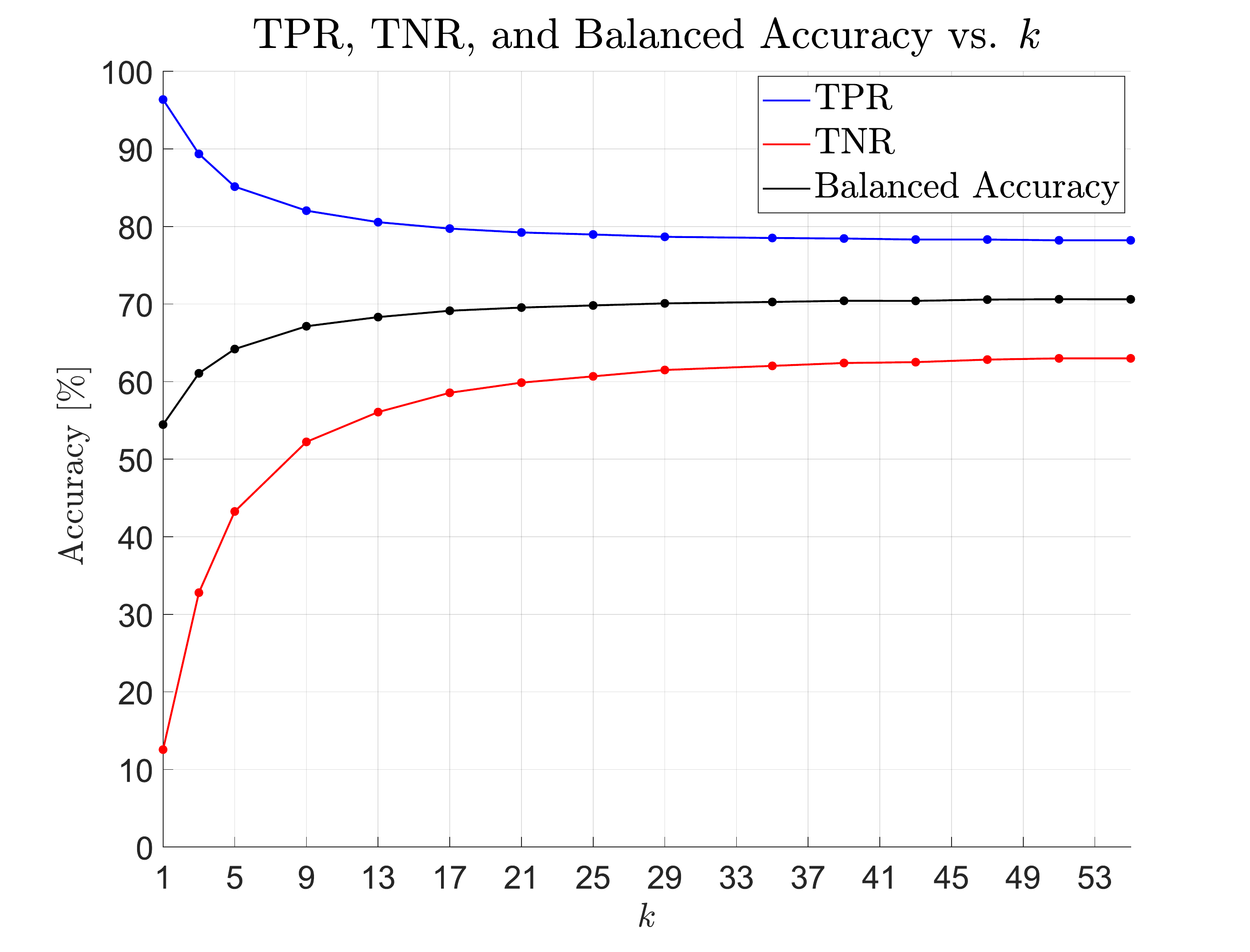}}
		\vspace{-0mm}	
	\end{subfigure}%
	\begin{subfigure}[]{.45\textwidth}
		\hspace*{0mm}\subcaptionbox{$k=13$\label{fig:487-ab}}
		{\includegraphics[scale=0.27]{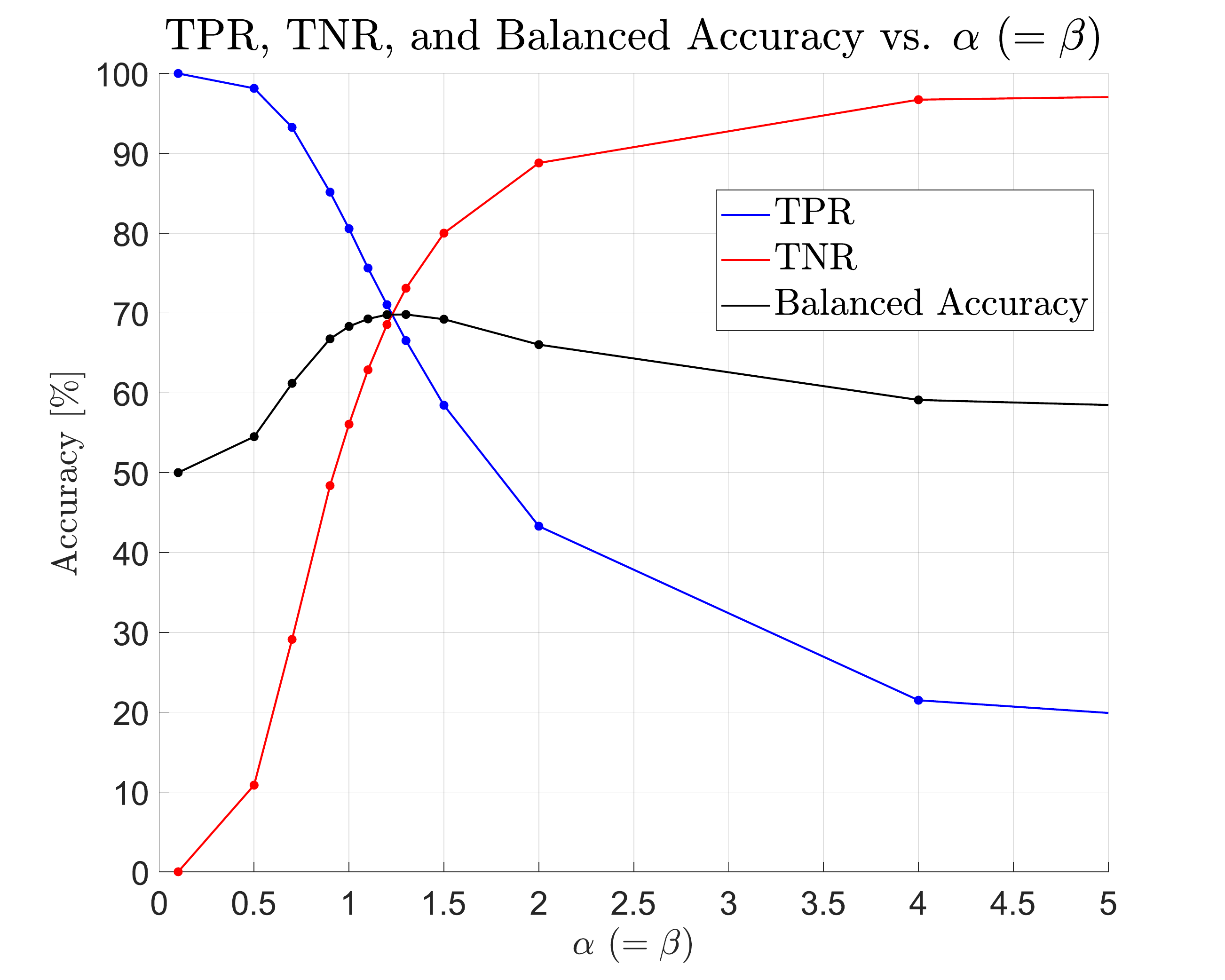}}
	\end{subfigure}
	\caption{Accuracy measures for the 487-astrobots swarm}
	\label{fig:487-k-ab}
\end{figure*}
\begin{figure*}
	\centering
	\hspace*{0mm}
	\begin{subfigure}[]{.45\textwidth}
		\hspace*{-0mm}\subcaptionbox{The variation of performance criteria in the case of the 487-astrobots scenario\label{fig:487-prf1}}
		{\includegraphics[scale=0.27]{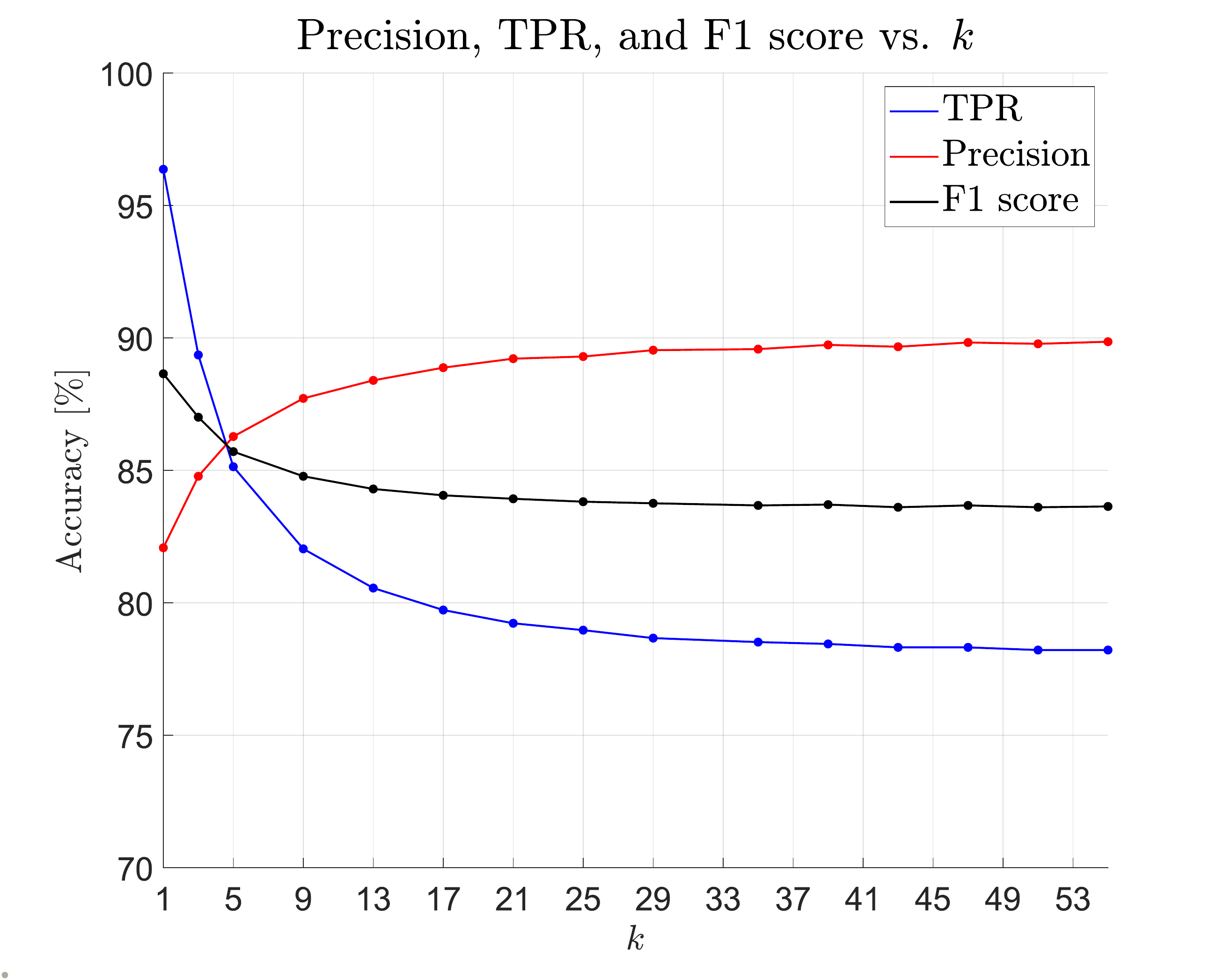}}
		\vspace{-0mm}	
	\end{subfigure}%
	\begin{subfigure}[]{.45\textwidth}
		\hspace*{0mm}\subcaptionbox{The comparative ROC curve evolution of both scenarios\label{fig:roc-comp}}
		{\includegraphics[scale=0.27]{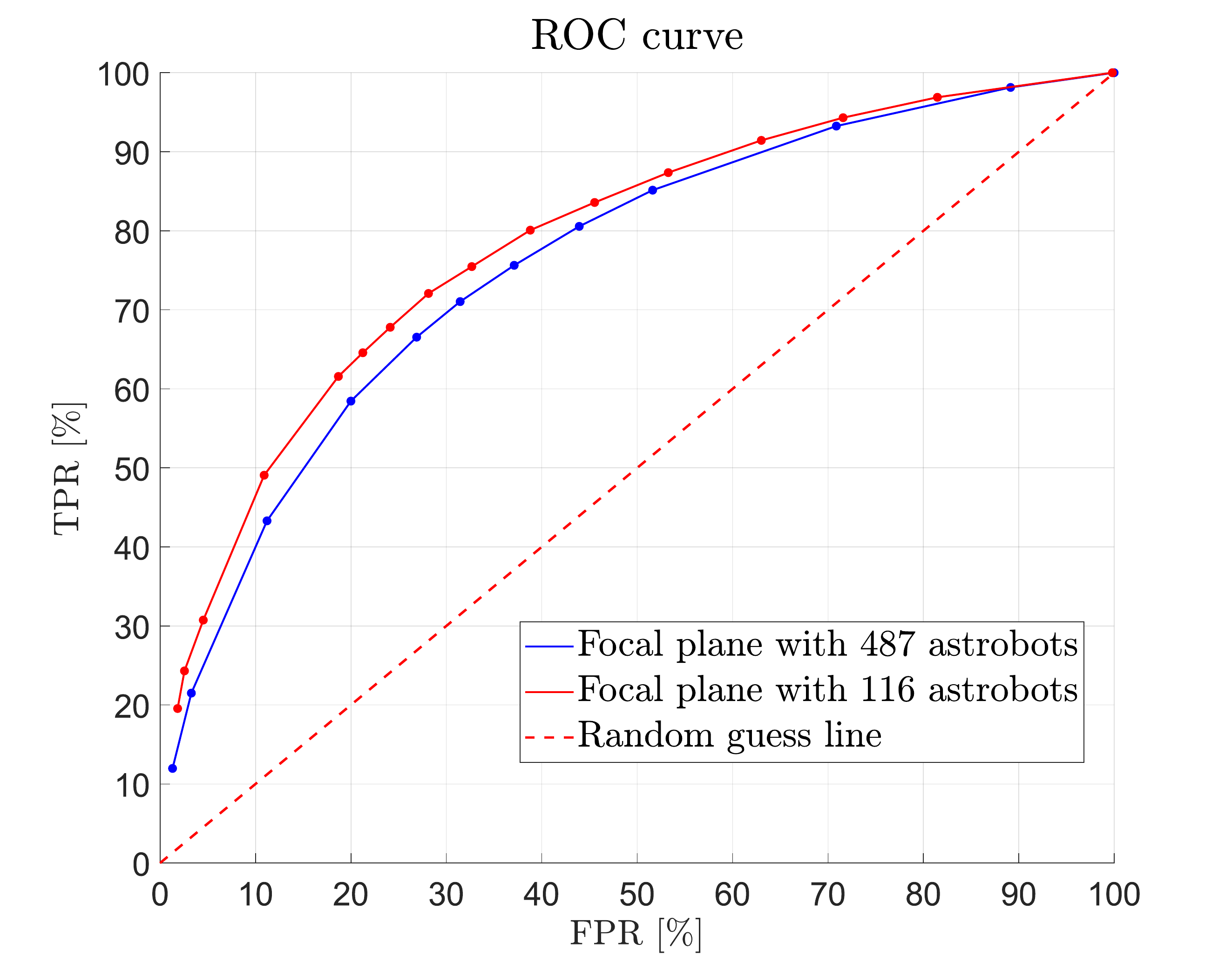}}
	\end{subfigure}
	\caption{Further accuracy results associated with both swarms}
	\label{fig:487-prf1-comp}
\end{figure*}

Confusion matrices of Fig. \ref{fig:487-con-neigh} reiterate the point that larger $k$ values give rise to the better predictions of the negatives. Fig. \ref{fig:487-neigh} witnesses the decrement of the balanced accuracy compared to the 116-astrobots swarm. The reason is that the 487-astrobots swarm comprises more total neighborhoods than the 116-astrobots swarm. The stability analysis of this case, similar to the previous case, also indicates the variations of the accuracy rates with respect to the hyperparameters as shown in Fig. \ref{fig:487-k-ab}. In particular, Fig. \ref{fig:487-k} exhibits that the algorithm is stable for $k>21$. Moreover, Fig. \ref{fig:487-prf1} illustrates the upper bound of the precision which is around $90\%$. Finally, we observe that the algorithm works on this 487-astrobots swarm almost as good as the 116-astrobots one. Namely, Fig. \ref{fig:487-prf1-comp} exhibits the ROC curve of the 487-astrobots case which is trivially closer to the random guess line compared to that of the 116-astrobots swarm. 
\section{Concluding remarks}
\label{sec:conc}
The first solution to the convergence prediction of populated packs of astrobots is studied. We observe that astrobot-to-target assignments provide a necessary feature subset of an astrobots swarm feature set to reach $\sim$80\% of accuracy in predicting the completely-converging set of the pairings. The $k$-NN nature of the proposed algorithm makes the metric design process intuitive enough to exploit the geometrical characteristics of astrobots and their neighborhoods. The presented strategy also enjoys a fairly restricted number of hyperparameters. So, the design process is not only relatively straightforward but tuning processes also require less computational resources.

This research indeed takes only necessary positional features of swarms to predicate convergences. However, it is imperative to look for extra features which obtain better accuracies such as parity, i.e., the motion direction of an astrobot. Needless to say that such feature expansion jeopardizes the computational efficiency of the prediction process as a trade-off. One may also utilize neural networks to train predictors which may noticeably provide noticeable accurate results. However, neural networks include many hyperparameters whose proper setting may be challenging specially if one would like to avoid computationally intensive grid searches.
\nocite{*}
\bibliographystyle{IEEEtran}
\bibliography{references}{}
\end{document}